\documentclass[runningheads]{llncs}
\usepackage{graphicx}
\usepackage{comment}
\usepackage[nosumlimits]{amsmath} %
\usepackage{amssymb} %
\usepackage{color}
\usepackage{enumitem}

\usepackage{floatrow}
\newfloatcommand{capbtabbox}{table}[][\FBwidth]

\usepackage{breqn}
\usepackage{mathtools}
\makeatletter
\newcommand{\subalign}[1]{%
  \vcenter{%
    \Let@ \restore@math@cr \default@tag
    \baselineskip\fontdimen10 \scriptfont\tw@
    \advance\baselineskip\fontdimen12 \scriptfont\tw@
    \lineskip\thr@@\fontdimen8 \scriptfont\thr@@
    \lineskiplimit\lineskip
    \ialign{\hfil$\m@th\scriptstyle##$&$\m@th\scriptstyle{}##$\hfil\crcr
      #1\crcr
    }%
  }%
}
\makeatother

\usepackage{tabularx}
\usepackage{multirow}
\usepackage{booktabs}   %

\newcolumntype{R}[1]{>{\raggedleft\let\newline\\\arraybackslash\hspace{0pt}}m{#1}}
\usepackage{threeparttable}
\usepackage{colortbl}

\usepackage{color}
\definecolor{Orange}{rgb}{0.9,0.5,0}
\definecolor{NavyBlue}{rgb}{0.1, 0.4, 0.8}
\definecolor{Magenta}{rgb}{0.8, 0.1, 0.6}

\usepackage{cleveref}
\Crefformat{figure}{#2Fig.~#1#3}
\Crefmultiformat{figure}{Figs.~#2#1#3}{ and~#2#1#3}{, #2#1#3}{ and~#2#1#3}
\newcommand{\bst}[1]{\textcolor[RGB]{27,158,119}{\mathbf{#1}}}
\newcommand{\bstt}[1]{\textcolor[RGB]{27,158,119}{\textbf{#1}}}

\usepackage[width=122mm,left=12mm,paperwidth=146mm,height=193mm,top=12mm,paperheight=217mm]{geometry}

\usepackage{capt-of,etoolbox}

\begin{document}
\pagestyle{headings}
\mainmatter

\title{GMM-UNIT: Unsupervised Multi-Domain and Multi-Modal Image-to-Image Translation via Attribute Gaussian Mixture Modeling}

\titlerunning{Abbreviated paper title}
\author{Yahui Liu\inst{1,2} \and
Marco De Nadai\inst{1} \and
Jian Yao\inst{3} \and
Nicu Sebe\inst{2} \and
Bruno Lepri\inst{1} \and
Xavier Alameda-Pineda\inst{4}}
\authorrunning{F. Author et al.}
\institute{FBK, Italy \and
University of Trento, Italy \and
Wuhan Unversity, China \and
Inria Grenoble Rh\^one-Alpes, France}

\maketitle

\begin{abstract}
Unsupervised image-to-image translation (UNIT) aims at learning a mapping between several visual domains by using unpaired training images. Recent studies have shown remarkable success for multiple domains but they suffer from two main limitations: they are either built from several \textit{two-domain mappings} that are required to be learned independently, or they generate low-diversity results, a problem known as \textit{mode collapse}. To overcome these limitations, we propose a method named GMM-UNIT, which is based on a content-attribute disentangled representation where the attribute space is fitted with a GMM. Each GMM component represents a domain, and this simple assumption has two prominent advantages. 
First, it can be easily extended to most multi-domain and multi-modal image-to-image translation tasks. 
Second, the continuous domain encoding allows for interpolation between domains and for extrapolation to unseen domains and translations. 
Additionally, we show how GMM-UNIT can be constrained down to different methods in the literature, meaning that GMM-UNIT is a unifying framework for unsupervised image-to-image translation.

\end{abstract}

\section{Introduction}
Translating images from one domain into another is a challenging task that has significant influence on many real-world applications where data are expensive, or impossible to obtain and to annotate. Image-to-Image translation models have indeed been used to increase the resolution of images~\cite{dong2014learning}, fill missing parts~\cite{pathak2016context}, transfer styles~\cite{gatys2016image}, synthesize new images from labels~\cite{liu2017unsupervised}, and help domain adaptation~\cite{bousmalis2017unsupervised,murez2018image}. 
In many of these scenarios, it is desirable to have a model mapping one image to multiple domains, while providing visual diversity (e.g.\ a day scene $\leftrightarrow$ night scene in different seasons). However, most of the existing models can either map an image to \emph{multiple} stochastic results in a single domain, or model \emph{multiple} domains in a deterministic fashion. In other words, the majority of the methods in the literature are either \emph{multi}-domain or \emph{multi}-modal.

Several reasons have hampered a stochastic translation of images to multiple domains. On the one hand, most of the Generative Adversarial Network (GAN) models assume a deterministic mapping~\cite{choi2018stargan,pumarola2018ganimation,zhu2017unpaired}, thus failing at modeling the correct distribution of the data~\cite{huang2018multimodal}. On the other hand, approaches based on Variational Auto-Encoders (VAEs) usually assume a shared and common zero-mean unit-variance normally distributed space~\cite{huang2018multimodal,zhu2017toward}, limiting to two-domain translations.

\begin{figure}[t]
\begin{center}
    \vspace{-2.2mm}
	\includegraphics[width=\textwidth]{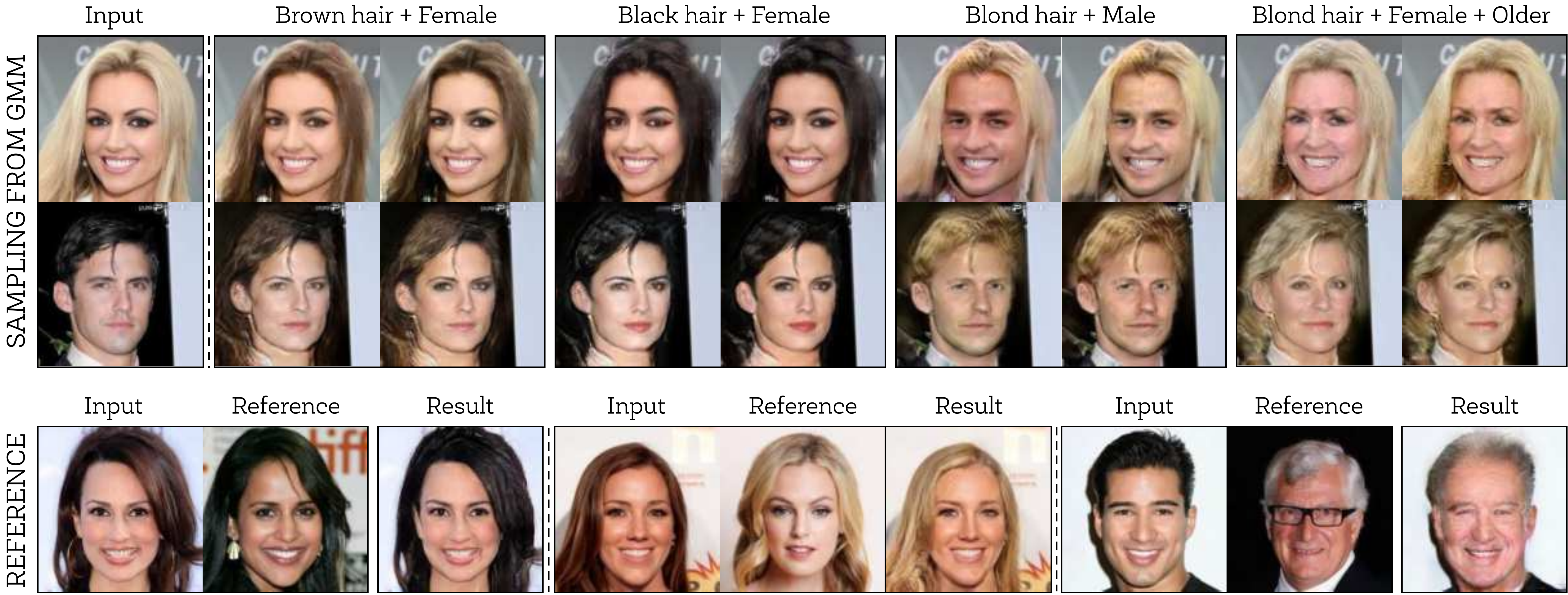}
	\captionof{figure}{GMM-UNIT is a multi-domain and multi-modal image-to-image translation model where the target domain can either be sampled from a distribution, or extracted from a reference image. The first two rows show diverse images generated for each domain translation. The last row shows translations from a reference image.}
	\label{fig:teaser}
    \vspace{-4.8mm}
\end{center}
\end{figure}

We propose a novel UNsupervised Image-to-image Translation (UNIT) model that disentangles the visual content from the domain attributes. The attribute latent space is assumed to follow a Gaussian Mixture Model (GMM), thus naming the method: GMM-UNIT (see \Cref{fig:teaser}). This simple assumption allows three key properties: \textit{mode-diversity} thanks to the stochastic nature of the probabilistic latent model, \textit{multi-domain translation} since the domains are represented as clusters in the same attribute spaces and \textit{few/zero-shot generation} since the continuity of the attribute representation allows interpolating between domains and extrapolating to unseen domains with very few or almost no observed data from these domains. The code and models will be made publicly available.

\section{Related work}
\vspace{-2mm}
Our work is best placed in the literature of image-to-image translation, where the challenge is to translate one image from a visual domain (e.g. summer) to another one (e.g. winter). This problem is inherently ill-posed, as there could be many mappings between two images. Thus, researchers tried to tackle the problem from different perspectives. The most impressive results on this task are undoubtedly related to GANs, which aim to synthesize new images as similar as possible to the real data through an adversarial approach between a Discriminator and a Generator. The former continuously learns to recognize real and fake images, while the latter tries to generate new images that are indistinguishable from the real data, and thus to fool the Discriminator. 
These networks can be effectively conditioned and thus generate new samples from a specific class \cite{chen2016infogan} and a latent vector extracted from the images. For example, \cite{isola2017image} and \cite{wang2018high} trained a conditional GAN to encode the latent features that are shared between images of the same domain and thus decode the features to images of the target domain in a one-to-one mapping. 
However, this approach is limited to supervised settings, where pairs of corresponding images in different domains are available (e.g. a photos-sketch image pair). In many cases, it is too expensive and unrealistic to collect a large amount of paired data. 

\noindent \textbf{Unsupervised Domain Translation}. Translating images from one domain to another without a paired supervision is particularly difficult, as the model has to learn how to represent both the content and the domain. Thus, constraints are needed to narrow down the space of feasible mappings between images.
\cite{taigmanPW17} proposed to minimize the feature-level distance between the generated and input images. \cite{liu2017unsupervised} created a shared latent space between the domains, which encourages different images to be mapped in the same latent space. \cite{zhu2017unpaired} proposed CycleGAN, which uses a cycle consistency loss that requires a generated image to be translated back to the original domain. Similarly, \cite{kim2017learning} used a reconstruction loss applying the same approach to both the target and input domains. \cite{mo2018instanceaware} later expanded the previous approach to the problem of translating multiple instances of objects in the same image. All these methods, however, are limited to a one-to-one domain mapping, thus requiring training multiple models for cross-domain translation. 
Recently, \cite{choi2018stargan} proposed StarGAN, a unified framework to translate images in a \textbf{multi-domain} setting through a single GAN model. To do so, they used a conditional label and a domain classifier ensuring network consistency when translating between domains. However, StarGAN is limited to a deterministic mapping between domains. 

\noindent \textbf{Style transfer}. A related problem is style transfer, which aims to transform the style of an image but not its content (e.g. from a photo to a Monet painting) to another image~\cite{donahue2018semantically,gatys2015neural,huang2017arbitrary,tenenbaum1997separating}. Differently from domain translation, usually the style is extracted from a single reference image. We will show that our model could be applied to style transfer as well.

\noindent \textbf{Multi-modal Domain Translation}. Most existing image-to-image translation methods are deterministic, thus limiting the diversity of the translated outputs. However, even in a one-to-one domain translation such as when we want to translate people's hair from blond to black, there could be multiple hair color shades that are not modeled in a deterministic mapping.
The straightforward solution would be injecting noise in the model, but it turned out to be worthless as GANs tend to ignore it~\cite{isola2017image,mathieu2015deep,zhu2017toward}. 
To address this problem, \cite{zhu2017toward} proposed BicycleGAN, which encourages the multi-modality in a paired setting through GANs and Variational Auto-Encoders (VAEs). \cite{almahairi2018augmented} have instead augmented CycleGAN with two latent variables for the input and target domains and showed that it is possible to increase diversity by marginalizing over these latent spaces.
\cite{huang2018multimodal} proposed MUNIT, which assumes that domains share a common content space but different style spaces. Then, they showed that by sampling from the style space and using Adaptive Instance Normalization (AdaIN)~\cite{huang2017arbitrary}, it is possible to have diverse and multimodal outputs. Similarly, \cite{ma2018exemplar} focused on the semantic consistency during the translation, and applied AdaIN to the feature-level space. Recently, \cite{MSGAN} proposed a mode seeking loss to encourage GANs to better explore the modes and help the network avoiding the mode collapse.

Altogether, the models in the literature are either multi-modal or multi-domain. Thus, one has to choose between generating diverse results and training one single model for multiple domains. Here, we propose a unified model to overcome this limitation. Concurrent to our work, DRIT++ \cite{lee2019drit++} also proposed a multi-modal and multi-domain model using a discrete domain encoding and assuming, however, a zero-mean unit-variance Gaussian shared space for multiple modes. We instead propose a content-attribute disentangled representation, where the attribute space fits a GMM distribution. A variational loss forces the latent representation to follow this GMM, where each component is associated to a domain. This is the key to provide for both multi-modal and multi-domain translation. In addition, GMM-UNIT is the first method proposing a continuous encoding of the domains, as opposed to the discrete encoding used in the literature. This is important because it allows for domain interpolation and extrapolation with very few or no data (few/zero-shot generation). The main properties of GMM-UNIT compared to the literature are shown in \Cref{table:models}.

\vspace{-1em}
\begin{table}[ht!]
    \caption{A comparison of the state of the art for image-to-image translation.  }
    \footnotesize
    \centering
    \setlength\tabcolsep{3.5pt}
    \begin{tabular}{@{}lcccl@{}}
        \toprule
        \textbf{Method} & \textbf{Unpaired} & \textbf{Multi-Domain} & \textbf{Multi-Modal} & \textbf{Domain encoding} \\
        \midrule
        CycleGAN~\cite{zhu2017unpaired} & \checkmark &  & & None \\
        BicycleGAN~\cite{zhu2017toward}   &   && \checkmark &  None \\
        MUNIT~\cite{huang2018multimodal}   & \checkmark && \checkmark & None \\
        StarGAN~\cite{choi2018stargan}   & \checkmark & \checkmark & & Discrete \\
        DRIT++~\cite{lee2019drit++} & \checkmark & \checkmark & \checkmark & Discrete \\
        GMM-UNIT & \checkmark & \checkmark & \checkmark & Continuous \\
         \bottomrule
    \end{tabular} \label{table:models}
\vspace{-3em}
\end{table}

\section{GMM-UNIT}
GMM-UNIT is an image-to-image translation model that translates an image from one domain to multiple domains in a stochastic fashion, which means that it generates multiple outputs with visual diversity for the same translation. 

Following recent seminal works~\cite{huang2018multimodal,lee2018diverse}, our model assumes that each image can be decomposed in a domain-invariant content space and a domain-specific attribute space. 
Given $Z$ attributes of a set of images, we model the attribute latent space through Gaussian Mixture Models (GMMs). Formally the probability density of the latent space $\mathbf{z}$ is defined as:
\begin{equation}
    p(\mathbf{z}) = \textstyle \sum_{k=1}^{K}\phi_k\mathcal{N}(\mathbf{z}; \pmb{\mu}^k, \pmb{\Sigma}^k)
    \label{eq:gmm}
\end{equation}
where $\mathbf{z}\in\mathbb{R}^Z$ denotes a random attribute vector sample, $\pmb{\mu}^k$ and $\pmb{\Sigma}^k$ denote respectively the mean vector and covariance matrix of the $k$-th GMM component, which is a $Z$-dimensional Gaussian ($\pmb{\mu}^k\in\mathbb{R}^Z$ and $\pmb{\Sigma}^k\in\mathbb{R}^{Z\times Z}$ is symmetric and positive definite). 
$\phi_k$ denotes the weight associated to the $k$-th component, where $\phi_k\geq 0$, $\sum_{k=1}^K \phi_k = 1$.
As later explained, in this paper we set $K=\left|\text{domains in the data}\right|$, which means that each Gaussian component represents a domain.
In other words, for an image $\mathbf{x}^n$ from domain $\pmb{\mathcal{X}}^n$ (i.e. $\mathbf{x} \sim p_{\pmb{\mathcal{X}}^n}$), then its latent attribute is assumed to follow $\mathbf{z}^n\sim{\cal N}(\pmb{\mu}^n, \pmb{\Sigma}^n)$, which is the $n$-th Gaussian component of the GMM that describes the domain $\pmb{\mathcal{X}}^n$. 

In the proposed representation, the domains are Gaussian components in a mixture. This simple yet effective model has one prominent advantage. Differently from previous works, where each domain is a category with a binary vector representation, we model the  distribution of attribute space.
The continuous encoding of the domains we here introduce allows us to navigate in the attribute latent space, thus generating images corresponding to domains that have never (or very little) been observed and allowing to interpolate between two domains.

We note that the state of the art models can be traced back particular case of GMMs. Existing multi-domain models such as StarGAN~\cite{choi2018stargan} or GANimation~\cite{pumarola2018ganimation} can be modeled with $K=\left|\text{domains in the data}\right|$ and  $ \forall k \ \pmb{\Sigma}^k=0$, thus only allowing the generation of a single result per domain translation.
Then, when $K=1$, $\pmb{\mu}=\textbf{0}$, and $\pmb{\Sigma}=\mathbf{I}$ it is possible to model the state of the art approaches in multi-modal translation~\cite{huang2018multimodal,zhu2017toward}, which share a unique latent space where every domain is overlapped, and it is thus necessary to train $K(K-1)$ models to achieve the multi-domain translation.
Finally, we can obtain DRIT++~\cite{lee2019drit++} by separating the attribute latent space into what they call an attribute space and a domain code. The former is a GMM with $K=1$, $\pmb{\mu}=\textbf{0}$, and $\pmb{\Sigma}=\mathbf{I}$, while the latter is another GMM with $K=|\text{domain in the data}|$ and  $ \forall k \ \pmb{\Sigma}^k=0$, which in \cite{lee2019drit++} is a one-hot encoding of the domain.
Thus, our GMM-UNIT is a generalization of the existing state of the art.
In the next sections, we formalize our model and show that the use of GMMs for the latent space allows learning multi-modal and multi-domain mappings, and also few/zero-shot image generation.

\begin{figure*}[h!t]
\centering
\includegraphics[width=0.92\textwidth]{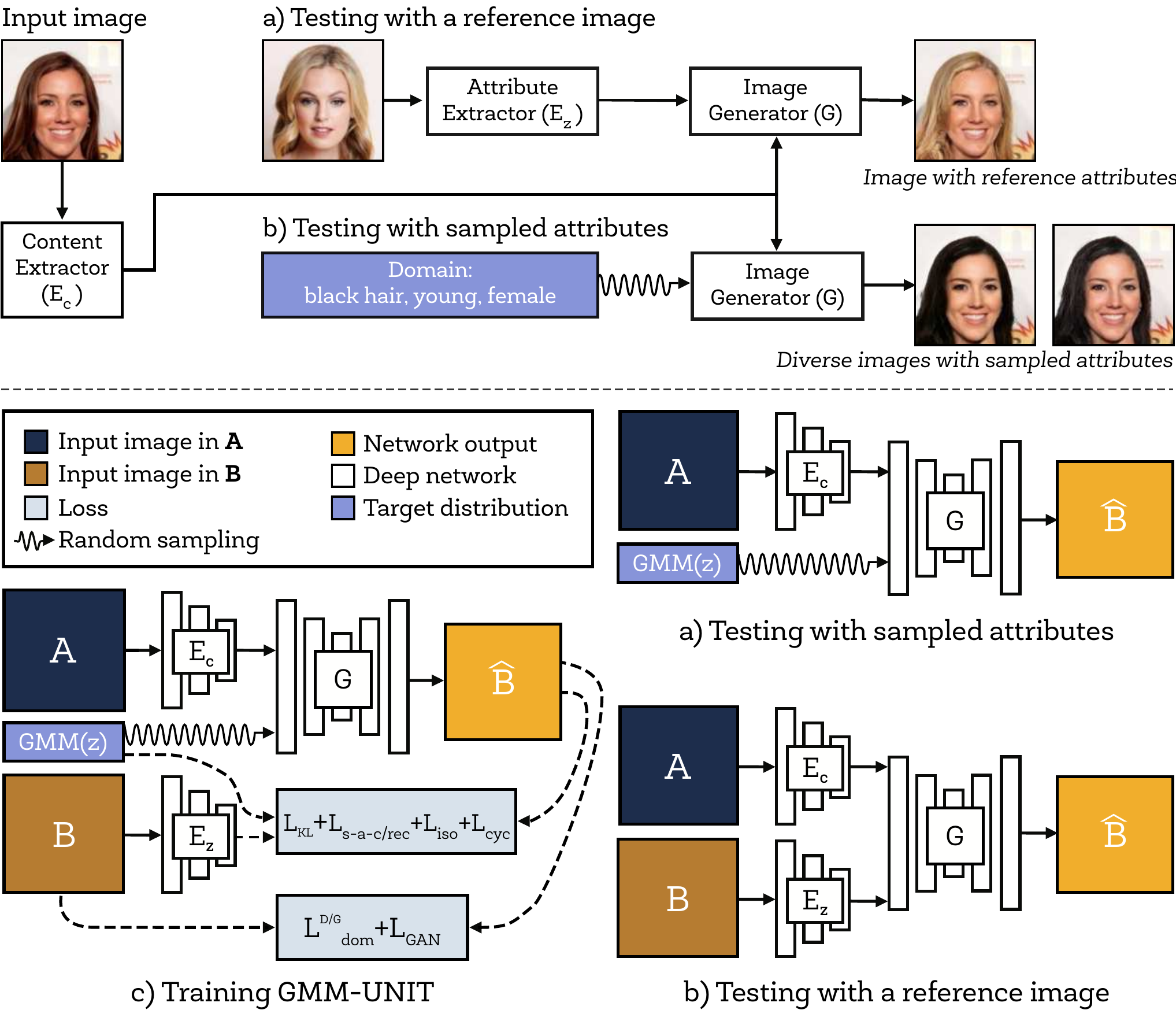}
\caption{GMM-UNIT translates an input image from one domain to a target domain. The content is extracted from the input image, while the attribute can be either sampled (a) or extracted from a reference image (b). In detail: c) Training phase to translate an image from domain $\pmb{A}$ to $\pmb{B}$. The generator uses the content of the input image (extracted by $E_c$) and the attribute of the target image (extracted by $E_z$) to generate an image in $\pmb{B}$. This image has the content of $\pmb{A}$ (e.g. Scarlett Johansson) but the attributes of $\pmb{B}$ (e.g. black hair). The attributes are modeled through a GMM. b) Testing phase where we use the content of an image in $\pmb{A}$ and the target attributes sampled from the GMM distribution of the attributes of domain $\pmb{B}$; c) Testing phase where we extract the content from an image in $\pmb{A}$ and the attributes from an image belonging to the target domain $\pmb{B}$. The style of this Figure is inspired from~\cite{zhu2017toward}.}
\label{fig:comparisons}
\end{figure*}

\subsection{The generative-discriminative approach}
GMM-UNIT follows the generative-discriminative philosophy. The generator inputs a \textit{content} latent code $\mathbf{c}\in\pmb{\mathcal{C}}\subset\mathbb{R}^C$ and an \textit{attribute} latent code $\mathbf{z}\in\pmb{\mathcal{Z}}\subset\mathbb{R}^Z$, and outputs a generated image $G(\mathbf{c},\mathbf{z})$. This image is then fed to a discriminator that must discern between ``real'' or ``fake'' images ($D_{\textrm{r/f}}$), and must also recognize the domain of the generated image ($D_{\textrm{dom}}$). 
The attribute and content latent representations need to be learned, and they are modeled by two architectures, namely a \textit{content extractor} $E_c$ and an \textit{attribute extractor} $E_z$. See \Cref{fig:comparisons} for a graphical representation of GMM-UNIT for an $\pmb{A}\leftrightarrow\pmb{B}$ domain translation.

In addition to tackling the problem of multi-domain and multi-modal translation, we would like these two extractors, content and attribute, to be \textit{disentangled}~\cite{huang2018multimodal}. 
This would constrain the learning and hopefully yield better domain translation, since the content would be as independent as possible from the attributes. 
We expect the attributes features to be related to the considered attributes, while the content features are supposed to be related to the rest of the image. Formally, the following two properties must hold:
\begin{description}
\item[Sampled attribute translation] \begin{multline*}
G(E_c(\mathbf{x}^m),\mathbf{z}^n)\sim p_{\pmb{\mathcal{X}}^n} \forall\; \mathbf{z}^n\sim {\cal N}(\pmb{\mu}^n, \pmb{\Sigma}^n),\; \mathbf{x}^m\sim p_{\pmb{\mathcal{X}}^m} ,\; n,m\in\{1,\ldots,K\}.\end{multline*}
\item[Extracted attribute translation] \begin{multline*} G(E_c(\mathbf{x}^m),E_z(\mathbf{x}^n))\sim p_{\pmb{\mathcal{X}}^n}\quad \forall \; \mathbf{x}^n\sim p_{\pmb{\mathcal{X}}^n},\; \mathbf{x}^m\sim p_{\pmb{\mathcal{X}}^m},\;n,m\in\{1,\ldots,K\}.\end{multline*}
\end{description}

\subsection{Training the GMM-UNIT}

The encoders $E_c$ and $E_z$, and the generator $G$ need to be learned to satisfy three main properties. \textbf{Consistency}: An image and its generated/extracted codes have to be consistent even after a translation from a domain $\pmb{A}$ to a domain $\pmb{B}$. \textbf{Fit}: The distribution of the attribute latent space must follow a GMM. \textbf{Realism}: The generated images must be indistinguishable from the real images. 
In the following, we discuss different losses used to force the overall pipeline to satisfy these properties.

In the \textbf{consistency} term, we include image, attribute and content reconstruction, as well as cycle consistency. More formally, we use the following losses:

\begin{itemize}[leftmargin=*, topsep=4pt]
\item \textit{Self-reconstruction} of any input image from its extracted content and attribute vectors:
\begin{equation*}
    \mathcal{L}_{\textrm{s/rec}} =  \textstyle \sum_{n=1}^K\mathbb{E}_{ \mathbf{x} \sim p_{\pmb{\mathcal{X}}^n}} \big[\|G(E_c(\mathbf{x}),E_z(\mathbf{x})) - \mathbf{x}\|_1 \big] 
    \label{eq:self}
\end{equation*}
\item \textit{Content reconstruction} from an image, translated into any domain:
\begin{dmath*}
    \mathcal{L}_{\textrm{c/rec}} = \textstyle \sum_{n,m=1}^K\mathbb{E}_{\mathbf{x}\sim p_{\pmb{\mathcal{X}}^n}, \mathbf{z}\sim {\cal N}(\pmb{\mu}^m, \pmb{\Sigma}^m)} \big[\| E_c(G(E_c(\mathbf{x}), \mathbf{z})) - E_c(\mathbf{x})\|_1 \big]
    \label{eq:cont}
\end{dmath*}

\item \textit{Attribute reconstruction} from an image translated with any content:
\begin{dmath*}
    \mathcal{L}_{\textrm{a/rec}} = \textstyle \sum_{n,m=1}^K\mathbb{E}_{\mathbf{x}\sim p_{\pmb{\mathcal{X}}^n}, \mathbf{z}\sim {\cal N}(\pmb{\mu}^m, \pmb{\Sigma}^m)} \big[\| E_z(G(E_c(\mathbf{x}), \mathbf{z})) - \mathbf{z}\|_1 \big]
    \label{eq:attr}
\end{dmath*}
\item \textit{Cycle consistency} when translating an image back to the original domain:
\begin{dmath*}
    \mathcal{L}_{\textrm{cyc}} = \textstyle \sum_{n,m=1}^K \mathbb{E}_{\mathbf{x}\sim p_{\pmb{\mathcal{X}}^n},\mathbf{z}\sim {\cal N}(\pmb{\mu}^m, \pmb{\Sigma}^m)}
    \big[\| G(E_c(G(E_c(\mathbf{x}), \mathbf{z})), E_z(\mathbf{x})) -\mathbf{x}  \|_1\big]
    \label{eq:cyc}
\end{dmath*}
\end{itemize}
We note that all these losses are used in prior work~\cite{choi2018stargan,huang2018multimodal,zhu2017unpaired,zhu2017toward} to constraint the infinite number of mappings that exist between an image in one domain and an image into another one. The $\mathcal{L}_1$ loss is used as it generates sharper results than the $\mathcal{L}_2$ loss~\cite{isola2017image}.
We also propose to complement the Attribute reconstruction with an isometry loss, to encourage the attribute extractor to be as similar as possible to the sampled attributes. Formally:
\begin{dmath*}
    \mathcal{L}_{\textrm{iso}} = \textstyle \sum_{n,m=1}^K\mathbb{E}_{\mathbf{x}\sim p_{\pmb{\mathcal{X}}^n}, \mathbf{z},\mathbf{z}'\sim {\cal N}(\pmb{\mu}^m, \pmb{\Sigma}^m)} \big[ |\| E_z(G(E_c(\mathbf{x}), \mathbf{z})) -  E_z(G(E_c(\mathbf{x}), \mathbf{z}'))\|_1 -  \|\mathbf{z}-\mathbf{z}' \|_1| \big]
\end{dmath*}

In the \textbf{fit} term we encourage both the attribute latent variable to follow the Gaussian mixture distribution and the generated images to follow the domain's distribution. We set two loss functions:

\begin{itemize}[leftmargin=*, topsep=4pt]
\item \textit{Kullback-Leibler divergence} between the extracted latent code and the model. Since the KL divergence between two GMMs is not analytically tractable, we resort on the fact that we know from which domain are we sampling and define:
    \begin{equation*}
        \mathcal{L}_{\textrm{KL}} = \textstyle \sum_{n=1}^K\mathbb{E}_{\mathbf{x}\sim p_{\pmb{\mathcal{X}}^n}} [\mathcal{D}_{KL}(E_z(\mathbf{x})\|\mathcal{N}(\pmb{\mu}^n, \pmb{\Sigma}^n))]
        \label{eq:kl}
    \end{equation*}
where $\mathcal{D}_{KL}(p\|q) = -\int p(t)\log\frac{p(t)}{q(t)}dt$ is the Kullback-Leibler divergence.  %

\item \textit{Domain classification} of generated and original images. For any given input image $\mathbf{x}$, we would like the method to classify it as its original domain, and to be able to generate from its content an image in any domain. Therefore, we need two different losses, one directly applied to the original images, and a second one applied to the generated images:
\begin{align*}
    \mathcal{L}_{\text{dom}}^D &= \textstyle \sum_{n=1}^K\mathbb{E}_{\mathbf{x}\sim p_{\pmb{\mathcal{X}}^n}, d_{\pmb{\mathcal{X}}^n}}[-\log D_{\text{dom}}(d_{\pmb{\mathcal{X}}^n}|\mathbf{x})],\\
    \mathcal{L}_{\text{dom}}^G &= \textstyle \sum_{n, m=1}^{K}\mathbb{E}_{\mathbf{x}\sim p_{\pmb{\mathcal{X}}^n},d_{\pmb{\mathcal{X}}^m}, \mathbf{z}\sim {\cal N}(\pmb{\mu}^m,\pmb{\Sigma}^m)
    }[-\log D_{\text{dom}}(d_{\pmb{\mathcal{X}}^m}|G(E_c(\mathbf{x}), \mathbf{z}))]
\end{align*}
\end{itemize}
where $d_{\pmb{\mathcal{X}}^n}$ is the label of domain $n$. Importantly, while the generator is trained using the second loss only, the discriminator $D_{\textrm{dom}}$ is trained using both.

The \textbf{realism} term tries to making the generated images indistinguishable from real images; we adopt the adversarial loss to optimize both the real/fake discriminator $D_{\textrm{r/f}}$ and the generator $G$:
\begin{dmath*}
    \mathcal{L}_{\text{GAN}} =  \textstyle \sum_{n,m=1}^K\mathbb{E}_{\mathbf{x}\sim p_{\pmb{\mathcal{X}}^n}}[-\log D_{\text{r/f}}(\mathbf{x})]  +  \mathbb{E}_{\subalign{\mathbf{x} \sim p_{\pmb{\mathcal{X}}^m}, \mathbf{z}\sim {\cal N}(\pmb{\mu}^n,\pmb{\Sigma}^n)}}[-\log(1-D_{\text{r/f}}(G(E_c(\mathbf{x}), \mathbf{z})))]
\end{dmath*}

\noindent The full objective function of our network is:
\begin{equation*}
\begin{split}
    \mathcal{L}_D =& \mathcal{L}_{\text{GAN}} + \mathcal{L}_{\text{dom}}^D\\
    \mathcal{L}_G =& \mathcal{L}_{\text{GAN}} + \lambda_{\text{s/rec}} \mathcal{L}_{\text{s/rec}} +  \mathcal{L}_{\text{c/rec}} + \mathcal{L}_{\text{a/rec}} \\ &+ \lambda_{\text{cyc}}\mathcal{L}_{\text{cyc}} +  \lambda_{\text{KL}}\mathcal{L}_{\text{KL}} + 
    \lambda_{\text{iso}}\mathcal{L}_{\text{iso}} + 
    \mathcal{L}_{\text{dom}}^G 
\end{split}
\end{equation*}
where $\{\lambda_{\textrm{s/rec}}, \lambda_{\textrm{cyc}}, \lambda_{\textrm{KL}}, \lambda_{\textrm{iso}}\}$ are hyper-parameters of weights for corresponding loss terms. The values of most of these parameters come from the literature. We refer to the Supplementary for the details.

\section{Experiments}
We perform extensive quantitative and qualitative analysis in three real-world tasks, namely: edges-shoes, digits and faces. First, we test GMM-UNIT on a simple task such as a one-to-one domain translation. Then, we move to the problem of multi-domain translation where each domain is independent from each other. Finally, we test our model on multi-domain translation where each domain is built upon different combinations of lower level attributes. Specifically, for this task, we test GMM-UNIT in a dataset containing over 40 labels related to facial attributes such as hair color, gender, and age. Each domain is then composed by combinations of these attributes, which might be mutually exclusive (e.g. either male or female) or mutually inclusive (e.g. blond and black hair).

Additionally, we show how the learned GMM latent space can be used to interpolate attributes and generate images in previously unseen domains.
Finally, we apply GMM-UNIT to the Style transfer task.

We compare our model to the state of the art of both multi-modal and multi-domain image translation problems. In the former, we select BicycleGAN~\cite{zhu2017toward}, MUNIT~\cite{zhu2017unpaired} and MSGAN~\cite{MSGAN}. In the latter, we compare with StarGAN~\cite{choi2018stargan} and DRIT++~\cite{lee2019drit++}, which is the only multi-modal and multi-domain method in the literature. However, StarGAN is not multi-modal. Thus, similarly to what done previously~\cite{zhu2017toward}, we modify StarGAN to be conditioned on Gaussian noise ($-0.2 \mathcal{N}(0,1) + 0.1$) in the input domain vector. We call this version of the model StarGAN* and we test it. More details are in the Supplementary.

\subsection{Metrics}
We quantitatively evaluate our method through image quality and diversity of generated images. The former is evaluated through the Fr\'echet Inception Distance (FID)~\cite{NIPS2017_7240}, while we evaluate the latter through the LPIPS~\cite{zhang2018unreasonable}. \\
\noindent\textbf{FID} We use FID to measure the distance between the generated and real distributions. Lower FID values indicate better quality of the generated images. We estimate the FID using 1000 input images and 10 samples per input v.s.\ randomly selected 10000 images from the target domain. \\
\noindent\textbf{LPIPS} The LPIPS distance is defined as the $\mathcal{L}_2$ distance between the features extracted by a deep learning model of two images. This distance has been demonstrated to match well the human perceptual similarity~\cite{zhang2018unreasonable}. Thus, following~\cite{huang2018multimodal,lee2018diverse,zhu2017toward}, we randomly select 100 input images and translate them to different domains. For each domain translation, we generate 10 images for each input image and evaluate the average LPIPS distance between the 10 generated images. Finally, we get the average of all distances. Higher LPIPS distance indicates better diversity among the generated images.

\subsection{Edges $\leftrightarrow$ Shoes: Two-domains Translation}
\begin{figure}[ht]
    \setlength{\tabcolsep}{1pt}
	\renewcommand{\arraystretch}{0.8}
    \newcommand{\sizea}{0.13\linewidth}
    \scriptsize
	\centering
	\begin{tabular}{c | ccccc | c}
	Input \& GT & StarGAN* & MUNIT & MSGAN & DRIT++ & GMM-UNIT & BicycleGAN \\
	   \includegraphics[width=\sizea]{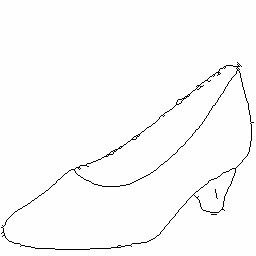} &
       \includegraphics[width=\sizea]{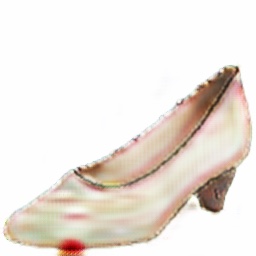} &
	   \includegraphics[width=\sizea]{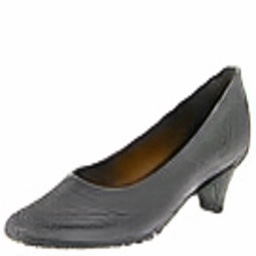} &
	   \includegraphics[width=\sizea]{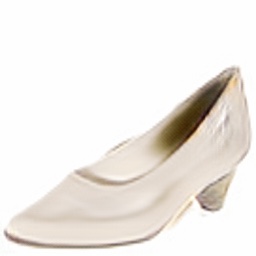} &
	   \includegraphics[width=\sizea]{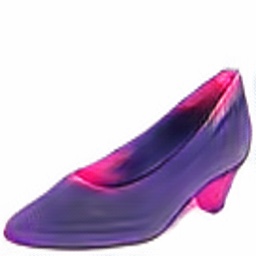} &
	   \includegraphics[width=\sizea]{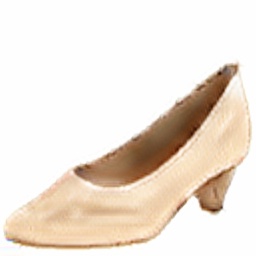} &
	   \includegraphics[width=\sizea]{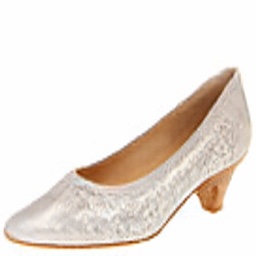}\\ 
	   \includegraphics[width=\sizea]{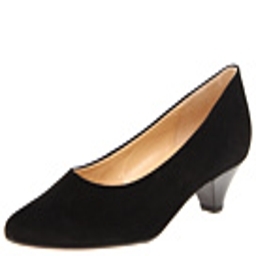} &
	   \includegraphics[width=\sizea]{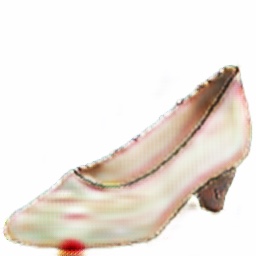} &
	   \includegraphics[width=\sizea]{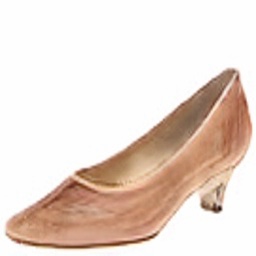} & 
	   \includegraphics[width=\sizea]{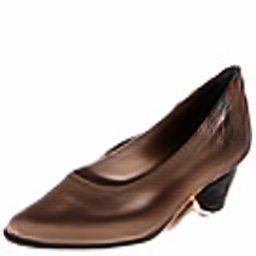} &
	   \includegraphics[width=\sizea]{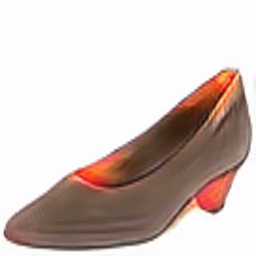} &
	   \includegraphics[width=\sizea]{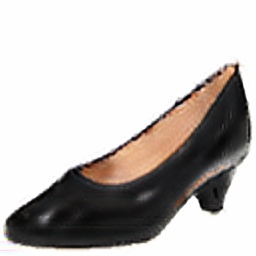} &
	   \includegraphics[width=\sizea]{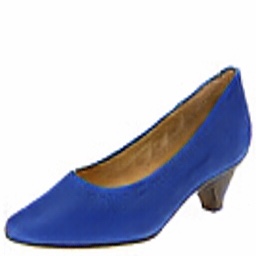}\\ 
	   &
	   \includegraphics[width=\sizea]{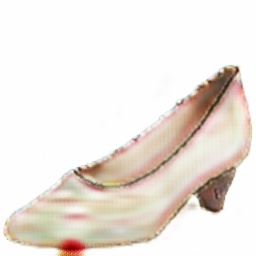} &
	   \includegraphics[width=\sizea]{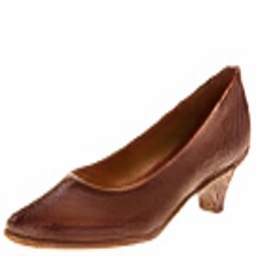} &
	   \includegraphics[width=\sizea]{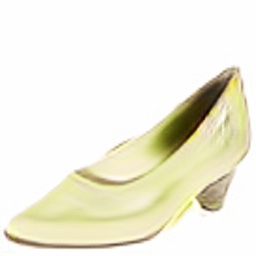} &
	   \includegraphics[width=\sizea]{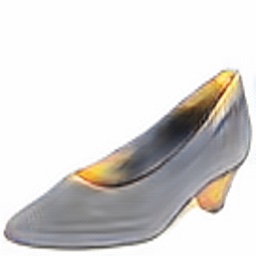} &
	   \includegraphics[width=\sizea]{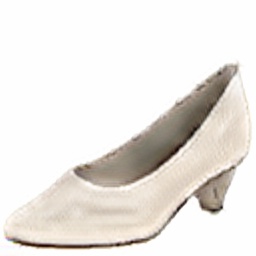} &
	   \includegraphics[width=\sizea]{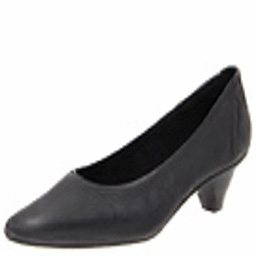}\\
	\end{tabular}
	\caption{Qualitative evaluation on the Edges $\to$ Shoes.\vspace{-4mm}}
	\label{Fig:edges2shoes}
\end{figure}

We first evaluate our model on a simpler task than multi-domain translation: two-domain translation (e.g. edges to shoes).
We use the dataset provided by~\cite{isola2017image,zhu2017unpaired} containing images of shoes and their edge maps generated by the Holistically-nested Edge Detection (HED)~\cite{xie2015holistically}. We resize all images to 256$\times$256 and train a single model for edges $\leftrightarrow$ shoes without using paired information. \Cref{Fig:edges2shoes} displays examples of shoes generated from the same sketch by all the state of the art models.
GMM-UNIT and MUNIT generate high-quality and diverse results that are almost indistinguishable from the ground truth and the results of BicycleGAN, which is a paired (supervised) method. Although, MSGAN and DRIT++ generate diverse images, they suffer from low quality results. The results of StarGAN* confirm the findings of previous studies that only adding noise does not increase diversity~\cite{isola2017image,mathieu2015deep,zhu2017toward}.
These results are confirmed in the quantitative evaluation displayed in Table~\ref{tab:QuantitativeResults_edges2shoes}. 
Our model generates images with high diversity and quality using half the parameters of the state of the art (MUNIT), which needs to be re-trained for each transformation. Particularly, the diversity is comparable to the paired model performance. These results show that this multi-modal and multi-domain model can be efficiently applied also to simpler tasks than multi-domain problems without much loss in performance, while other multi-domain models suffer in this setting.
We refer to the Supplementary for additional results on this task.

\begin{table*}[ht]
    \caption{Quantitative evaluation on the Edges $\to$ Shoes dataset. The best performance for unpaired (unsupervised) models is in \bstt{green}. $^\dag$ refers to supervised method. MM and MD stands for Multi-Modal and Multi-Domain respectively.}
    \setlength{\tabcolsep}{4pt}
    \footnotesize
	\centering
	\begin{tabular}{@{}l ccc @{\quad} rr @{\quad} r @{\quad} r@{}}
	\toprule
	\textbf{Model} & \textbf{Unpaired} & \textbf{MM} & \textbf{MD} & \textbf{FID}$\downarrow$ & \textbf{LPIPS}$\uparrow$ & \textbf{Params$\downarrow$} 
	\\
	\midrule
	StarGAN*~\cite{choi2018stargan} & \checkmark && \checkmark & 140.41 & $.002\pm .000$ & $53.23M\times1$ \\
	MUNIT~\cite{huang2018multimodal} & \checkmark & \checkmark & & $\bst{54.52}$ & $\bst{.227\pm .001}$ & $23.52M\times2$ \\
	MSGAN~\cite{MSGAN} & \checkmark &\checkmark & & 111.19 & $.221\pm .003$ & $65.03M\times2$ \\
	DRIT++~\cite{lee2019drit++} & \checkmark &\checkmark & \checkmark & 123.87 & $.233\pm .002$ & $54.06M\times1$ \\
	GMM-UNIT & \checkmark &\checkmark & \checkmark & 58.46 & $.200 \pm .002$ &  $\bst{23.52M \times 1}$\\\midrule
	BicycleGAN$^\dag$~\cite{zhu2017toward} && \checkmark & & 47.43 & $.199\pm .001$ & $64.30M\times2$ \\  
	\bottomrule
	\end{tabular}
	\label{tab:QuantitativeResults_edges2shoes}
\end{table*}

\subsection{Digits: Single-attribute Multi-domain Translation}
We then increase the complexity of the task by evaluating our model in a multi-domain translation setting, where each domain is composed by digits collected in different scenes. We use the Digits-Five dataset introduced in~\cite{xu2018deep}, from which we select three different domains, namely MNIST~\cite{lecun1998gradient}, MNIST-M~\cite{ganin2014unsupervised}, and Street View House Numbers (SVHN)~\cite{yuval2011reading}. During the training, given that all images are resized to 32$\times$32, we reduce the depth of our model and compared models. 
We compare our model with the state-of-the-art on multi-domain translation, and we show in \Cref{tab:QuantitativeResults_digits_ce} the quantitative results. We add in the Supplementary extensive qualitative results for space limit reasons.

From these results we conclude that StarGAN* fails at generating diversity, while GMM-UNIT generates images with higher quality and diversity than all the state-of-the-art models.
Additional experiments carried out implementing a StarGAN*-like GMM-UNIT (i.e. setting $\pmb{\Sigma}^k=0, \;\forall k$) indeed produced similar results. Specifically, the StarGAN*-like GMM-UNIT tends to generate for each input image one single (deterministic) output and thus the corresponding LPIPS scores are zero.
We refer to the Supplementary for additional results on this task.

\begin{table*}[ht]
    \caption{Quantitative evaluation on the Digits and Faces datasets. The best performance is in \bstt{green}. For Faces, we also evaluate the diversity on the background.}
    \setlength{\tabcolsep}{2pt}
    \footnotesize
	\centering
	\begin{tabular}{@{}l c cc c rr c rrr@{}}
	\toprule
	\multirow{2}{*}{\textbf{Model}} & \phantom{i} & \multirow{2}{*}{\textbf{MM}} & \multirow{2}{*}{\textbf{MD}} & \phantom{i} & \multicolumn{2}{c}{\textbf{Digits}} & \phantom{i} & \multicolumn{3}{c}{\textbf{Faces}} 
	\\
	\cmidrule(l{2pt}){6-7} \cmidrule{9-11} 
	&&&&& FID$\downarrow$ & LPIPS$\uparrow$ && FID$\downarrow$ & LPIPS$\uparrow$ & LPIPS$_b$$\downarrow$ 
	\\
	\midrule
	StarGAN*~\cite{choi2018stargan} && & \checkmark && 69.11 & $.006\pm.000$ && 51.68 & $.002\pm .000$ &  $.035\pm .010$ \\
	DRIT++~\cite{lee2019drit++} && \checkmark & \checkmark && 88.94 & $.058\pm.001$ && 55.64 & $.017\pm.001$ & $.055\pm .001$\\
	GMM-UNIT && \checkmark & \checkmark && $\bst{60.43}$ & $\bst{.124\pm .002}$ && $\bst{46.21}$ & $\bst{.048 \pm .002}$  & $\bst{.022\pm .004}$ \\
	\bottomrule
	\end{tabular}
	\label{tab:QuantitativeResults_digits_ce}
\end{table*}

\subsection{Faces: Multi-attribute Multi-domain Translation}
We also evaluate GMM-UNIT in the complex setting of multi-domain translation in a dataset of facial attributes.
We use the Celebfaces Attributes (CelebA) dataset~\cite{liu2015deep}, which contains 202,599 face images of celebrities where each face is annotated with 40 binary attributes. We apply central cropping to the initial 178$\times$218 size
images to 178$\times$178, then resize the cropped images to 128$\times$128. We randomly select 2,000 images for testing and use all remaining images for training. 
This dataset is composed of some attributes that are mutually exclusive (e.g. either male or female) and those that are mutually inclusive (e.g. people could have both blond and black hair).
Thus, we model each attribute as a different GMM component. For this reason, we can generate new images for all the combinations of attributes by sampling from the GMM. 
As aforementioned, this is not possible for state-of-the-art models such as StarGAN and DRIT++, as they use one-hot domain codes to represent the domains. 
To be consistent with the state of the art (StarGAN) we show five binary attributes: hair color (\textit{black, blond, brown}),
gender (\textit{male/female}), and age (\textit{young/old}). These five attributes allow GMM-UNIT to generate 32 domains.

We observed that image-to-image translation is sensitive to complex background information. In fact, models are inclined to manipulate the intensity and details of pixels that are not related to the desired attribute transformation. 
Hence, we add a convolutional layer at the end the of decoder $G$ to learn a one-channel attention mask $\mathbf{M}$ in an unsupervised manner. Hence, the final prediction $\widehat{\mathbf{B}}$ is obtained through combining the input image $\mathbf{A}$ and its initial prediction $\widetilde{\mathbf{B}}$ through: $\widehat{\mathbf{B}} = \widetilde{\mathbf{B}} \cdot \mathbf{M} + \mathbf{A} \cdot (1-\mathbf{M}) $.
We also apply the attention layer to Edges $\leftrightarrow$ Shoes and Digits, but find that it provides no noticeable improvements in the results. 

\begin{figure}[ht]
    \setlength{\tabcolsep}{1pt}
	\renewcommand{\arraystretch}{0.8}
    \newcommand{\sizea}{0.13\linewidth}
    \scriptsize
	\centering
	\begin{tabular}{c|ccccc}
	Input & Black hair & Brown hair & Blond hair & Blond+Male
	& Blond+Older \\
	   \includegraphics[width=\sizea]{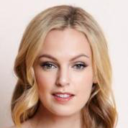} &
       \includegraphics[width=\sizea]{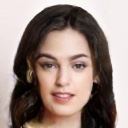} &
	   \includegraphics[width=\sizea]{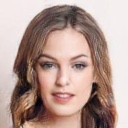} &
	   \includegraphics[width=\sizea]{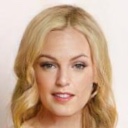} &
	   \includegraphics[width=\sizea]{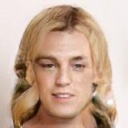} &
	   \includegraphics[width=\sizea]{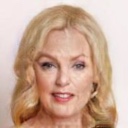} \\ &
	   \includegraphics[width=\sizea]{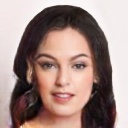} &
	   \includegraphics[width=\sizea]{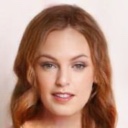} & 
	   \includegraphics[width=\sizea]{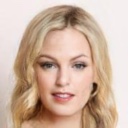} &
	   \includegraphics[width=\sizea]{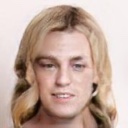} &
	   \includegraphics[width=\sizea]{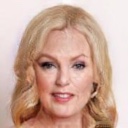} \\ &
	   \includegraphics[width=\sizea]{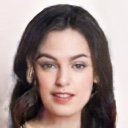} &
	   \includegraphics[width=\sizea]{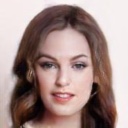} &
	   \includegraphics[width=\sizea]{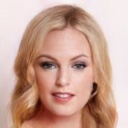} &
	   \includegraphics[width=\sizea]{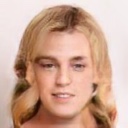} &
	   \includegraphics[width=\sizea]{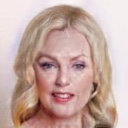} \\
	\end{tabular}
	\caption{Facial expression synthesis results on the CelebA dataset with different attribute combinations. Each row represents a different output sampled from the model.}
	\label{Fig:celeba}
\end{figure}

\Cref{Fig:celeba} shows some generated results of our model. We can see that GMM-UNIT learns to translate images to simple attributes such as blond hair, but also to translate images with combinations of them (e.g. blond hair and male). Moreover, we can see that the rows show different realizations of the model thus demonstrating the stochastic approach of GMM-UNIT.
These results are corroborated by \Cref{tab:QuantitativeResults_digits_ce} that shows that our model is superior to StarGAN* and DRIT++ in both quality and diversity of generated images. 
Particularly, the use of an attention mechanism allows our model to achieve diversity only on the part of the image that is involved in the transformation (e.g. hair and face for gender and hair translation). To demonstrate this, we compute the LPIPS distance between the background of the input image and the generated images (LPIPS$_b$). \Cref{tab:QuantitativeResults_digits_ce} that our model is the best at preserving the original background information. In \Cref{Fig:diversity} we show the difference between the diversity we achieve and DRIT++ diversity. GMM-UNIT preserves the background while it changes the face and create diverse hair styles, while DRIT++ just changes the overall color intensity and affects parts of the image not related to the attributes, which is not desirable.
Extensive results are displayed in the Supplementary. 

\vspace{-6mm}

\vspace{-6mm}
\subsection{Style transfer}
\begin{figure}[ht]
	\renewcommand{\tabcolsep}{1pt}
	\renewcommand{\arraystretch}{0.8}
	\newcommand{\sizea}{0.13\linewidth}
	\centering
	\begin{tabular}{c|ccccc} 
	\includegraphics[width=\sizea]{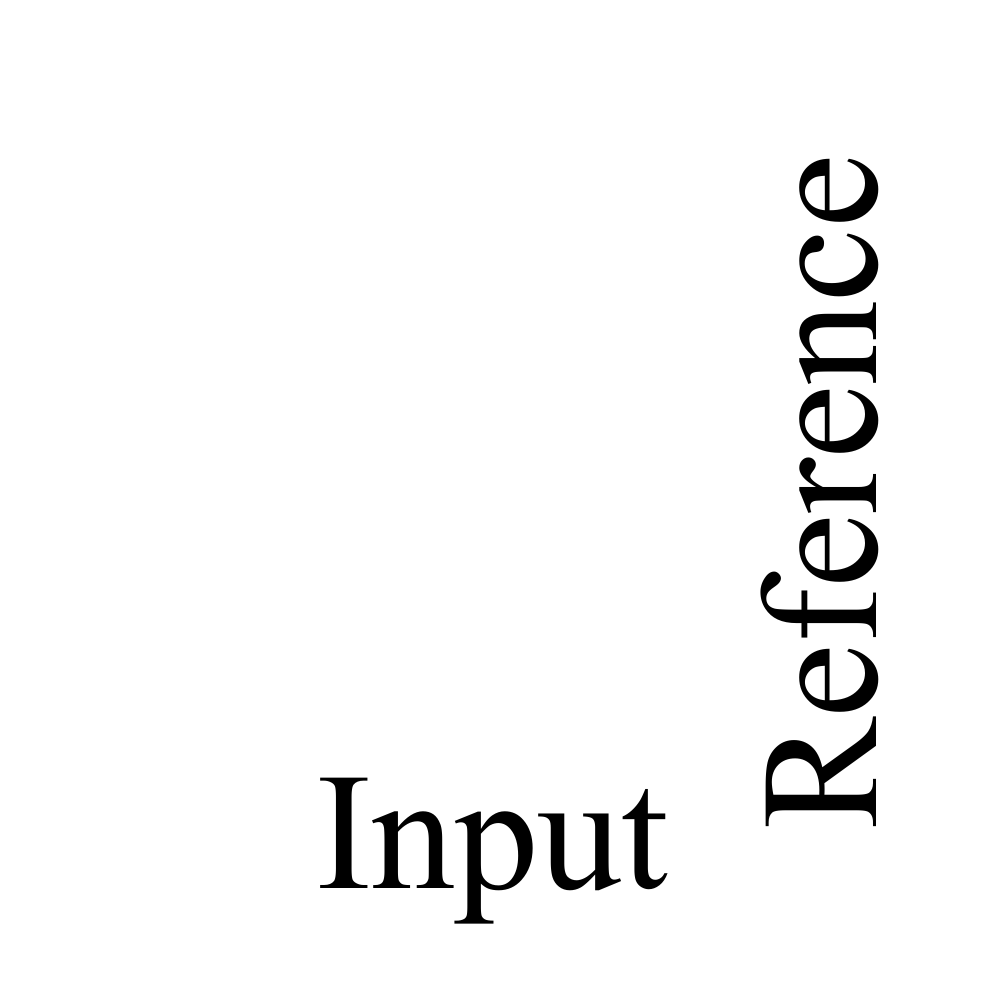} &%
	\includegraphics[width=\sizea]{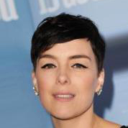} &%
	\includegraphics[width=\sizea]{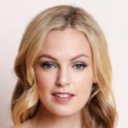} &%
	\includegraphics[width=\sizea]{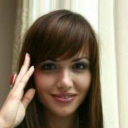} &%
	\includegraphics[width=\sizea]{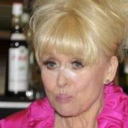} &
	\includegraphics[width=\sizea]{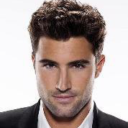} \\
	\hline
	\includegraphics[width=\sizea]{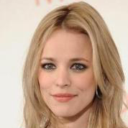} & 
	\includegraphics[width=\sizea]{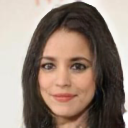} &
	 \includegraphics[width=\sizea]{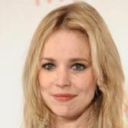} & 
	\includegraphics[width=\sizea]{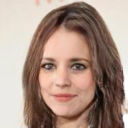} &
	\includegraphics[width=\sizea]{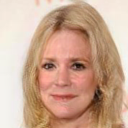} & 
	\includegraphics[width=\sizea]{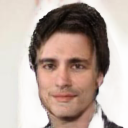} \\
	\includegraphics[width=\sizea]{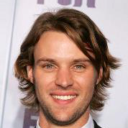} & 
	\includegraphics[width=\sizea]{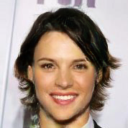} &
	 \includegraphics[width=\sizea]{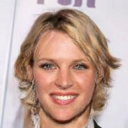} & 
	\includegraphics[width=\sizea]{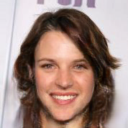} &
	\includegraphics[width=\sizea]{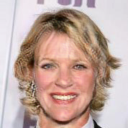} & 
	\includegraphics[width=\sizea]{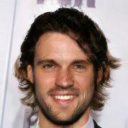} \\
	\end{tabular}
	\caption{Examples of GMM-UNIT applied on the Style transfer task. The style is here extracted from a single reference images provided by the user.}
	\label{Fig:samples2}
\end{figure}

We evaluate our model on style transfer, which is a specific task where the style is usually extracted from a single reference image. 
Thus, we randomly select two input images and synthesize new images where, instead of sampling from the GMM distribution, we extract the style (through $E_z$) from some reference images.
\Cref{Fig:samples2} shows that the generated images are sharp and realistic, showing that our method can also be effectively applied to Style transfer.

\begin{figure*}[h!t]	
	\renewcommand{\tabcolsep}{1pt}
	\renewcommand{\arraystretch}{0.8}
	\newcommand{\sizea}{0.11\linewidth}
	\centering
	\footnotesize
	\begin{tabular}{c|ccc|ccc}
	        Input & \multicolumn{3}{c}{Black+Blond+Female+Young} & \multicolumn{3}{c}{Black+Blond+Male+Young} \\
		\includegraphics[width=\sizea]{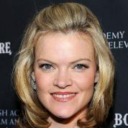} &
		\includegraphics[width=\sizea]{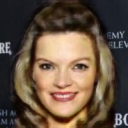} &
		\includegraphics[width=\sizea]{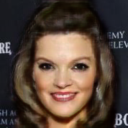} &
		\includegraphics[width=\sizea]{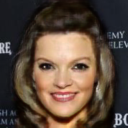} &
		\includegraphics[width=\sizea]{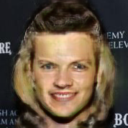} &
		\includegraphics[width=\sizea]{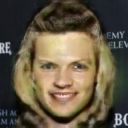} &
		\includegraphics[width=\sizea]{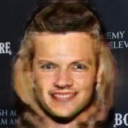}\\
	\end{tabular}
    \caption{Generated images in previously unseen combinations of attributes.\label{Fig:unseen}}
\end{figure*}

\begin{figure*}[h!t]	
	\renewcommand{\tabcolsep}{1pt}
	\renewcommand{\arraystretch}{0.8}
	\newcommand{\sizea}{0.10\linewidth}
	\centering
	\footnotesize
	\begin{tabular}{c|cccccccc}
	        Input & \multicolumn{3}{l}{Black hair+Female+Young} & \multicolumn{2}{c}{$\longleftrightarrow$}  & \multicolumn{3}{r}{Blond hair+Female+Young}\\
		\includegraphics[width=\sizea]{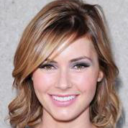} &
		\includegraphics[width=\sizea]{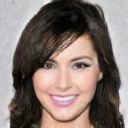} &
		\includegraphics[width=\sizea]{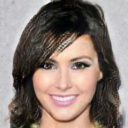} &
		\includegraphics[width=\sizea]{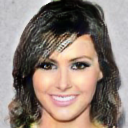} &
		\includegraphics[width=\sizea]{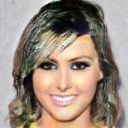} &
		\includegraphics[width=\sizea]{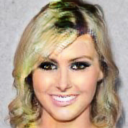} & 
		\includegraphics[width=\sizea]{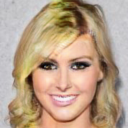} & 
		\includegraphics[width=\sizea]{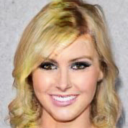} &   
		\includegraphics[width=\sizea]{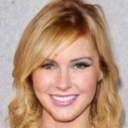} \\
	\end{tabular}
	\caption{An example of domain interpolation given an input image.}
	\label{Fig:interpolation}
\end{figure*}

\subsection{Domain interpolation and extrapolation}
In addition, we evaluate the ability of GMM-UNIT to synthesize new images with attributes that are extremely scarce or non present in the training dataset. To do so, we select three combinations of attributes consisting of less than two images in the CelebA dataset: \textit{Black hair+Blond hair+Male+Young} and \textit{Black hair+Blond hair+Female+Young}.

\Cref{Fig:unseen} shows that learning the continuous and multi-modal latent distribution of attributes allow to effectively generate images as zero- or few-shot generation. At the best of our knowledge, we are the first ones being able to translate images in previously unseen domains at no additional cost. Recent literature on zero-pair translation learning indeed scale linearly with the number of domains~\cite{wang2018mix}. This ability can be of vital importance in tasks where labels are extremely imbalanced.

Finally, we show that by learning  the full latent distribution of the attributes we can do attribute interpolation both intra- and inter-domains. In contrast, state of the art methods such as \cite{lee2019drit++} can only do intra-domain interpolations due to their discrete domain encoding. Other works such as Chen \emph{et al.}~\cite{chen2019homomorphic} are focused on explicitly learning an interpolation and use a reference image to do the same task, while we can either interpolate between two reference images or between any two points in the attribute latent space (by sampling these points/vectors), even for multiple attributes.
\Cref{Fig:interpolation} shows some generated images through a linear interpolation between two given attributes, while in Supplementary we show that we can also do intra-domain interpolations.

\subsection{Ablation study}
\begin{figure*}
\begin{floatrow}
\ffigbox{%
  \renewcommand{\tabcolsep}{1pt}
	\renewcommand{\arraystretch}{0.8}
	\newcommand{\sizea}{0.23\linewidth}
	\centering
	\scriptsize
	\begin{tabular}{c|ccc}
	        Input & \multicolumn{3}{c}{Black hair + Female} \\
	        \cmidrule{2-4} & \multicolumn{2}{c}{GMM-UNIT} & DRIT++
	        \\
		\includegraphics[width=\sizea]{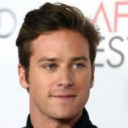} &
		\includegraphics[width=\sizea]{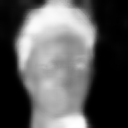} &
		\includegraphics[width=\sizea]{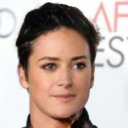} &
		\includegraphics[width=\sizea]{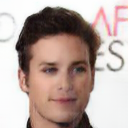} 
		\\
		&
		\includegraphics[width=\sizea]{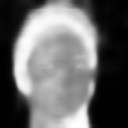} &
		\includegraphics[width=\sizea]{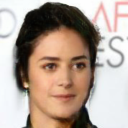} &
		\includegraphics[width=\sizea]{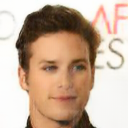} \\
	\end{tabular}
}{%
	\caption{GMM-UNIT diversity is only on the subject thanks to the attention, while DRIT++ changes also the background.}
	\label{Fig:diversity}
}
\capbtabbox{%
    \footnotesize
    \centering
  \begin{tabular}{@{}l rrrr@{}}
	\toprule
		\textbf{Model} & \textbf{FID$\downarrow$}  & \textbf{LPIPS$\uparrow$} \\
		\midrule
		GMM-UNIT (A) &  $\bst{60.43}$ &  $.124\pm .002$  \\
	    (A) w/o $\mathcal{L}_{\textrm{cyc}}$ & 84.06 & $\bst{.138\pm .003}$ \\
		(A) w/o $\mathcal{L}_{\textrm{a/rec}}$ & 62.20 & $.121\pm .002$ \\
	    (A) w/o $\mathcal{L}_{\textrm{iso}}$ & 63.70 & $.115\pm .002$ \\
	    (A) w/o \emph{disent.} & 60.72  & $.097\pm .003$  \\
		\bottomrule
	\end{tabular}
}{%
  \caption{Ablation study performance on the Digits dataset.}
  \label{tab:ablation_digits_details}
}
\end{floatrow}
\end{figure*}

Given that the importance of $\mathcal{L}_{s/rec}$ and $\mathcal{L}_{dom}$ was verified in previous works (i.e.\ CycleGAN and StarGAN), and that $\mathcal{L}_{c/rec}, \mathcal{L}_{KL}$ are necessary to the model convergence, we compare GMM-UNIT with three variants of the model that ablate $\mathcal{L}_{\textrm{cyc}}$, $\mathcal{L}_{\textrm{a/rec}}$ and $\mathcal{L}_{\textrm{iso}}$ in the Digits dataset.
\Cref{tab:ablation_digits_details} shows the results of the ablation.
As expected, $\mathcal{L}_{\textrm{cyc}}$ is needed to have higher image quality, and we observe that it increases the diversity because of noisy results. When $\mathcal{L}_{\textrm{a/rec}}$ is removed image quality decreases, but $\mathcal{L}_{\textrm{iso}}$ still helps to learn the attributes space. Finally, without $\mathcal{L}_{\textrm{iso}}$ we observe that both diversity and quality decrease, thus confirming the need of all these losses.
For the first time from its introduction in \cite{huang2018multimodal}, we also test for the \emph{disentangled} assumption of visual content and attributes.
Although we cannot test the network removing the attribute extractor $E_z$, we remove the content extractor $E_c$ and change the generator $G$ to have $\mathbf{x}$ and $\mathbf{z}$ as input.
We observe that the results are similar, although the diversity decreases substantially. This means that the disentanglement approach needs to be further studied in the multiple architectures and tasks that propose it~\cite{gonzalez2018image,huang2018multimodal,wu2019transgaga} to understand its necessity and contribution. 
We refer to Supplementary for the disentanglement and the additional ablation results broken down by domain.

\vspace{-2mm}
\section{Conclusion}
In this paper, we present a novel image-to-image translation model that maps images to multiple domains and provides a stochastic translation. GMM-UNIT disentangles the content of an image from its attributes and represents the attribute space with a GMM, which allows us to have a continuous encoding of domains. This has two main advantages: first, it can easily be extended to most multi-domain and multi-modal image-to-image translation tasks.  Second, GMM-UNIT allows for interpolation across-domains and the translation of images into previously unseen domains.

We conduct extensive experiments in three different tasks, namely two-domain translation, multi-domain translation and multi-attribute multi-domain translation. We show that GMM-UNIT achieves quality and diversity superior to state of the art, most of the times with fewer parameters.
Future work includes the possibility to thoroughly learn the mean vectors of the GMM from the data and extending the experiments to a higher number of GMM components per domain.

\bibliographystyle{splncs04}
\bibliography{egbib}

\begin{thebibliography}{10}
\providecommand{\url}[1]{\texttt{#1}}
\providecommand{\urlprefix}{URL }
\providecommand{\doi}[1]{https://doi.org/#1}

\bibitem{almahairi2018augmented}
Almahairi, A., Rajeswar, S., Sordoni, A., Bachman, P., Courville, A.: Augmented
  cyclegan: Learning many-to-many mappings from unpaired data. In: ICML (2018)

\bibitem{ba2016layer}
Ba, J.L., Kiros, J.R., Hinton, G.E.: Layer normalization. arXiv preprint
  arXiv:1607.06450  (2016)

\bibitem{bousmalis2017unsupervised}
Bousmalis, K., Silberman, N., Dohan, D., Erhan, D., Krishnan, D.: Unsupervised
  pixel-level domain adaptation with generative adversarial networks. In: CVPR.
  pp. 3722--3731 (2017)

\bibitem{chen2016infogan}
Chen, X., Duan, Y., Houthooft, R., Schulman, J., Sutskever, I., Abbeel, P.:
  Infogan: Interpretable representation learning by information maximizing
  generative adversarial nets. In: NIPS. pp. 2172--2180 (2016)

\bibitem{chen2019homomorphic}
Chen, Y.C., Xu, X., Tian, Z., Jia, J.: Homomorphic latent space interpolation
  for unpaired image-to-image translation. In: CVPR. pp. 2408--2416 (2019)

\bibitem{choi2018stargan}
Choi, Y., Choi, M., Kim, M., Ha, J.W., Kim, S., Choo, J.: Stargan: Unified
  generative adversarial networks for multi-domain image-to-image translation.
  In: CVPR. pp. 8789--8797 (2018)

\bibitem{donahue2018semantically}
Donahue, C., Balsubramani, A., McAuley, J., Lipton, Z.C.: Semantically
  decomposing the latent spaces of generative adversarial networks. In: ICLR
  (2018)

\bibitem{dong2014learning}
Dong, C., Loy, C.C., He, K., Tang, X.: Learning a deep convolutional network
  for image super-resolution. In: ECCV. pp. 184--199. Springer (2014)

\bibitem{ganin2014unsupervised}
Ganin, Y., Lempitsky, V.: Unsupervised domain adaptation by backpropagation.
  In: ICML (2014)

\bibitem{gatys2015neural}
Gatys, L.A., Ecker, A.S., Bethge, M.: A neural algorithm of artistic style.
  arXiv preprint arXiv:1508.06576  (2015)

\bibitem{gatys2016image}
Gatys, L.A., Ecker, A.S., Bethge, M.: Image style transfer using convolutional
  neural networks. In: CVPR. pp. 2414--2423 (2016)

\bibitem{gonzalez2018image}
Gonzalez-Garcia, A., van~de Weijer, J., Bengio, Y.: Image-to-image translation
  for cross-domain disentanglement. In: NIPS. pp. 1287--1298 (2018)

\bibitem{NIPS2017_7240}
Heusel, M., Ramsauer, H., Unterthiner, T., Nessler, B., Hochreiter, S.: Gans
  trained by a two time-scale update rule converge to a local nash equilibrium.
  In: NIPS (2017)

\bibitem{huang2017arbitrary}
Huang, X., Belongie, S.: Arbitrary style transfer in real-time with adaptive
  instance normalization. In: ICCV. pp. 1501--1510 (2017)

\bibitem{huang2018multimodal}
Huang, X., Liu, M.Y., Belongie, S., Kautz, J.: Multimodal unsupervised
  image-to-image translation. In: ECCV. pp. 172--189 (2018)

\bibitem{isola2017image}
Isola, P., Zhu, J.Y., Zhou, T., Efros, A.A.: Image-to-image translation with
  conditional adversarial networks. In: CVPR. pp. 1125--1134 (2017)

\bibitem{kim2017learning}
Kim, T., Cha, M., Kim, H., Lee, J.K., Kim, J.: Learning to discover
  cross-domain relations with generative adversarial networks. In: ICML. pp.
  1857--1865. JMLR. org (2017)

\bibitem{kingma2014adam}
Kingma, D.P., Ba, J.: Adam: A method for stochastic optimization. In: ICLR
  (2015)

\bibitem{lecun1998gradient}
LeCun, Y., Bottou, L., Bengio, Y., Haffner, P., et~al.: Gradient-based learning
  applied to document recognition. Proceedings of the IEEE  \textbf{86}(11),
  2278--2324 (1998)

\bibitem{lee2018diverse}
Lee, H.Y., Tseng, H.Y., Huang, J.B., Singh, M., Yang, M.H.: Diverse
  image-to-image translation via disentangled representations. In: ECCV. pp.
  35--51 (2018)

\bibitem{lee2019drit++}
Lee, H.Y., Tseng, H.Y., Mao, Q., Huang, J.B., Lu, Y.D., Singh, M., Yang, M.H.:
  Drit++: Diverse image-to-image translation via disentangled representations.
  International Journal of Computer Vision  (2020).
  \doi{10.1007/s11263-019-01284-z},
  \url{https://doi.org/10.1007/s11263-019-01284-z}

\bibitem{liu2017unsupervised}
Liu, M.Y., Breuel, T., Kautz, J.: Unsupervised image-to-image translation
  networks. In: NIPS. pp. 700--708 (2017)

\bibitem{liu2015deep}
Liu, Z., Luo, P., Wang, X., Tang, X.: Deep learning face attributes in the
  wild. In: ICCV. pp. 3730--3738 (2015)

\bibitem{ma2018exemplar}
Ma, L., Jia, X., Georgoulis, S., Tuytelaars, T., Gool, L.V.: Exemplar guided
  unsupervised image-to-image translation with semantic consistency. In: ICLR
  (2019)

\bibitem{maaten2008visualizing}
Maaten, L.v.d., Hinton, G.: Visualizing data using t-sne. Journal of machine
  learning research  \textbf{9}(Nov),  2579--2605 (2008)

\bibitem{MSGAN}
Mao, Q., Lee, H.Y., Tseng, H.Y., Ma, S., Yang, M.H.: Mode seeking generative
  adversarial networks for diverse image synthesis. In: CVPR (2019)

\bibitem{mathieu2015deep}
Mathieu, M., Couprie, C., LeCun, Y.: Deep multi-scale video prediction beyond
  mean square error. arXiv preprint arXiv:1511.05440  (2015)

\bibitem{mo2018instanceaware}
Mo, S., Cho, M., Shin, J.: Instance-aware image-to-image translation. In: ICLR
  (2019)

\bibitem{murez2018image}
Murez, Z., Kolouri, S., Kriegman, D., Ramamoorthi, R., Kim, K.: Image to image
  translation for domain adaptation. In: CVPR. pp. 4500--4509 (2018)

\bibitem{yuval2011reading}
Netzer, Y., Wang, T., Coates, A., Bissacco, A., Wu, B., Ng, A.Y.: Reading
  digits in natural images with unsupervised feature learning. In: NIPS
  Workshop on Deep Learning and Unsupervised Feature Learning (2011)

\bibitem{pathak2016context}
Pathak, D., Krahenbuhl, P., Donahue, J., Darrell, T., Efros, A.A.: Context
  encoders: Feature learning by inpainting. In: CVPR. pp. 2536--2544 (2016)

\bibitem{pumarola2018ganimation}
Pumarola, A., Agudo, A., Martinez, A.M., Sanfeliu, A., Moreno-Noguer, F.:
  Ganimation: Anatomically-aware facial animation from a single image. In:
  ECCV. pp. 818--833 (2018)

\bibitem{selvaraju2017grad}
Selvaraju, R.R., Cogswell, M., Das, A., Vedantam, R., Parikh, D., Batra, D.:
  Grad-cam: Visual explanations from deep networks via gradient-based
  localization. In: Proceedings of the IEEE international conference on
  computer vision. pp. 618--626 (2017)

\bibitem{simonyan2014very}
Simonyan, K., Zisserman, A.: Very deep convolutional networks for large-scale
  image recognition. arXiv preprint arXiv:1409.1556  (2014)

\bibitem{taigmanPW17}
Taigman, Y., Polyak, A., Wolf, L.: Unsupervised cross-domain image generation.
  In: ICLR (2017)

\bibitem{tenenbaum1997separating}
Tenenbaum, J.B., Freeman, W.T.: Separating style and content. In: NIPS. pp.
  662--668 (1997)

\bibitem{ulyanov2017improved}
Ulyanov, D., Vedaldi, A., Lempitsky, V.: Improved texture networks: Maximizing
  quality and diversity in feed-forward stylization and texture synthesis. In:
  CVPR. pp. 6924--6932 (2017)

\bibitem{wang2018high}
Wang, T.C., Liu, M.Y., Zhu, J.Y., Tao, A., Kautz, J., Catanzaro, B.:
  High-resolution image synthesis and semantic manipulation with conditional
  gans. In: CVPR. pp. 8798--8807 (2018)

\bibitem{wang2018mix}
Wang, Y., van~de Weijer, J., Herranz, L.: Mix and match networks:
  encoder-decoder alignment for zero-pair image translation. In: CVPR. pp.
  5467--5476 (2018)

\bibitem{wu2019transgaga}
Wu, W., Cao, K., Li, C., Qian, C., Loy, C.C.: Transgaga: Geometry-aware
  unsupervised image-to-image translation. In: CVPR. pp. 8012--8021 (2019)

\bibitem{xie2015holistically}
Xie, S., Tu, Z.: Holistically-nested edge detection. In: ICCV. pp. 1395--1403
  (2015)

\bibitem{xu2015empirical}
Xu, B., Wang, N., Chen, T., Li, M.: Empirical evaluation of rectified
  activations in convolutional network. arXiv preprint arXiv:1505.00853  (2015)

\bibitem{xu2018deep}
Xu, R., Chen, Z., Zuo, W., Yan, J., Lin, L.: Deep cocktail network:
  Multi-source unsupervised domain adaptation with category shift. In: CVPR.
  pp. 3964--3973 (2018)

\bibitem{zhang2018unreasonable}
Zhang, R., Isola, P., Efros, A.A., Shechtman, E., Wang, O.: The unreasonable
  effectiveness of deep features as a perceptual metric. In: CVPR. pp. 586--595
  (2018)

\bibitem{zhou2016learning}
Zhou, B., Khosla, A., Lapedriza, A., Oliva, A., Torralba, A.: Learning deep
  features for discriminative localization. In: Proceedings of the IEEE
  conference on computer vision and pattern recognition. pp. 2921--2929 (2016)

\bibitem{zhu2017unpaired}
Zhu, J.Y., Park, T., Isola, P., Efros, A.A.: Unpaired image-to-image
  translation using cycle-consistent adversarial networks. In: ICCV. pp.
  2223--2232 (2017)

\bibitem{zhu2017toward}
Zhu, J.Y., Zhang, R., Pathak, D., Darrell, T., Efros, A.A., Wang, O.,
  Shechtman, E.: Toward multimodal image-to-image translation. In: NIPS. pp.
  465--476 (2017)

\end{thebibliography}

\appendix 
\section{Implementation details}
\label{suppl:implementation}

Our deep neural model architecture is built upon the state-of-the-art methods MUNIT~\cite{huang2018multimodal}, BicycleGAN~\cite{zhu2017toward} and StarGAN~\cite{choi2018stargan}. 
As shown in \Cref{tab:architecture}, we apply Instance Normalization (IN)~\cite{ulyanov2017improved}
to the content encoder $E_c$, while we apply Adaptive Instance Normalization (AdaIN)~\cite{huang2017arbitrary} and Layer Normalization (LN)~\cite{ba2016layer} for the decoder $G$. For the discriminator network, we use Leaky ReLU~\cite{xu2015empirical} with a negative slope of 0.2. We note that we reduce the number of layers of the discriminator on the Digits dataset.

\begin{table}[!ht]
	\centering
	\scriptsize
	\caption{GMM-UNIT network architecture. We use the following notations: $Z$: the dimension of attribute vector, $n$: the number of attributes, N: the number of output channels, K: kernel size, S: stride size, P: padding size, CONV: a convolutional layer, GAP: a global average pooling layer, UPCONV: a 2$\times$bilinear upsampling layer followed by a convolutional layer, FC: fully connected layer. We set $C=1$ in Edges2shoes and Digits, $C=8$ in Faces. $\dag$ refers to be optional.}
	\begin{tabular}{@{}lll@{}}
	    \toprule
	    \textbf{Part} & \textbf{Input} $\rightarrow$ \textbf{Output Shape} & \textbf{Layer Information} \\ \midrule
		\multirow{7}{*}{$E_c$} & ($h$, $w$, 3) $\rightarrow$ ($h$, $w$, 64) & CONV-(N64, K7x7, S1, P3), IN, ReLU \\
		& ($h$, $w$, 64) $\rightarrow$ ($\frac{h}{2}$, $\frac{w}{2}$, 128) & CONV-(N128, K4x4, S2, P1), IN, ReLU \\
		& ($\frac{h}{2}$, $\frac{w}{2}$, 128) $\rightarrow$ ($\frac{h}{4}$, $\frac{w}{4}$, 256) & CONV-(N256, K4x4, S2, P1), IN, ReLU \\ 
		& ($\frac{h}{4}$, $\frac{w}{4}$, 256) $\rightarrow$ ($\frac{h}{4}$, $\frac{w}{4}$, 256) & Residual Block: CONV-(N256, K3x3, S1, P1), IN, ReLU \\ 
		& ($\frac{h}{4}$, $\frac{w}{4}$, 256) $\rightarrow$ ($\frac{h}{4}$, $\frac{w}{4}$, 256) & Residual Block: CONV-(N256, K3x3, S1, P1), IN, ReLU \\ 
		& ($\frac{h}{4}$, $\frac{w}{4}$, 256) $\rightarrow$ ($\frac{h}{4}$, $\frac{w}{4}$, 256) & Residual Block: CONV-(N256, K3x3, S1, P1), IN, ReLU \\ 
		& ($\frac{h}{4}$, $\frac{w}{4}$, 256) $\rightarrow$ ($\frac{h}{4}$, $\frac{w}{4}$, 256) & Residual Block: CONV-(N256, K3x3, S1, P1), IN, ReLU \\ \midrule
		\multirow{8}{*}{$E_z$} & ($h$, $w$, 3) $\rightarrow$ ($h$, $w$, 64) & CONV-(N64, K7x7, S1, P3), ReLU \\
		& ($h$, $w$, 64) $\rightarrow$ ($\frac{h}{2}$, $\frac{w}{2}$, 128) & CONV-(N128, K4x4, S2, P1), ReLU \\
		& ($\frac{h}{2}$, $\frac{w}{2}$, 128) $\rightarrow$ ($\frac{h}{4}$, $\frac{w}{4}$, 256) & CONV-(N256, K4x4, S2, P1), ReLU \\ 
		& ($\frac{h}{4}$, $\frac{w}{4}$, 256) $\rightarrow$ ($\frac{h}{8}$, $\frac{w}{8}$, 256) & CONV-(N256, K4x4, S2, P1), ReLU \\
		& ($\frac{h}{8}$, $\frac{w}{8}$, 256) $\rightarrow$ ($\frac{h}{16}$, $\frac{w}{16}$, 256) & CONV-(N256, K4x4, S2, P1), ReLU \\
		& ($\frac{h}{8}$, $\frac{w}{8}$, 256) $\rightarrow$ (1, 1, 256) & GAP \\ \cmidrule{2-3}
		& (256,) $\rightarrow$ ($CZ$,)  & FC-(N$CZ$) \\
		& (256,) $\rightarrow$ ($CZ$,)  & FC-(N$CZ$) \\
		\midrule
		\multirow{7}{*}{$G$} & ($\frac{h}{4}$, $\frac{w}{4}$, 256) $\rightarrow$ ($\frac{h}{4}$, $\frac{w}{4}$, 256) & Residual Block: CONV-(N256, K3x3, S1, P1), AdaIN, ReLU \\
		& ($\frac{h}{4}$, $\frac{w}{4}$, 256) $\rightarrow$ ($\frac{h}{4}$, $\frac{w}{4}$, 256) & Residual Block: CONV-(N256, K3x3, S1, P1), AdaIN, ReLU \\
		& ($\frac{h}{4}$, $\frac{w}{4}$, 256) $\rightarrow$ ($\frac{h}{4}$, $\frac{w}{4}$, 256) & Residual Block: CONV-(N256, K3x3, S1, P1), AdaIN, ReLU \\
		& ($\frac{h}{4}$, $\frac{w}{4}$, 256) $\rightarrow$ ($\frac{h}{4}$, $\frac{w}{4}$, 256) & Residual Block: CONV-(N256, K3x3, S1, P1), AdaIN, ReLU \\
		& ($\frac{h}{4}$, $\frac{w}{4}$, 256) $\rightarrow$ ($\frac{h}{2}$, $\frac{w}{2}$, 128) & UPCONV-(N128, K5x5, S1, P2), LN, ReLU \\ 
		& ($\frac{h}{2}$, $\frac{w}{2}$, 128) $\rightarrow$ ($h$, $w$, 64) & UPCONV-(N64, K5x5, S1, P2), LN, ReLU \\ 
		& (${h}$, ${w}$, 64) $\rightarrow$ (${h}$, ${w}$, 3) & CONV-(N3, K7x7, S1, P3), Tanh \\ 
		& \dag(${h}$, ${w}$, 64(+1)) $\rightarrow$ (${h}$, ${w}$, 1) & CONV-(N3, K7x7, S1, P3), Sigmoid \\ \midrule
		\multirow{6}{*}{$D$} & ($h$, $w$, 3) $\rightarrow$ ($\frac{h}{2}$, $\frac{w}{2}$, 64) & CONV-(N64, K4x4, S2, P1), Leaky ReLU \\
		& ($\frac{h}{2}$, $\frac{w}{2}$, 64) $\rightarrow$ ($\frac{h}{4}$, $\frac{w}{4}$, 128) & CONV-(N128, K4x4, S2, P1), Leaky ReLU \\
		& ($\frac{h}{4}$, $\frac{w}{4}$, 128) $\rightarrow$ ($\frac{h}{8}$, $\frac{w}{8}$, 256) & CONV-(N256, K4x4, S2, P1), Leaky ReLU \\
		& ($\frac{h}{8}$, $\frac{w}{8}$, 256) $\rightarrow$ ($\frac{h}{16}$, $\frac{w}{16}$, 512) & CONV-(N512, K4x4, S2, P1), Leaky ReLU \\  \cmidrule{2-3}
		& ($\frac{h}{16}$, $\frac{w}{16}$, 512) $\rightarrow$ ($\frac{h}{16}$, $\frac{w}{16}$, 1) & CONV-(N1, K1x1, S1, P0) \\ 
		& ($\frac{h}{16}$, $\frac{w}{16}$, 512) $\rightarrow$ (1, 1, $n$) & CONV-(N$n$, K$\frac{h}{16}$x$\frac{w}{16}$, S1, P0) \\ 
		\bottomrule
	\end{tabular}
	\label{tab:architecture}
\end{table}

We use the Adam optimizer~\cite{kingma2014adam} with $\beta_1$ = 0.5, $\beta_2$ = 0.999, and an initial learning rate of 0.0001. The learning rate is decreased by half every 2e5 iterations. In all experiments, we use a batch size of 1 for Edges2shoes and Faces and batch size of 32 for Digits. And we set the loss weights to $\lambda_{\text{s/rec}}$ = 10, $\lambda_{\text{cyc}}$ = 10, $\lambda_{\text{KL}}$ = 0.1, and $\lambda_{\text{iso}}$ = 0.1. We use the domain-invariant perceptual loss with weight 0.1 in all experiments. Random mirroring is applied during training.

\subsection{GMM}
While the GMM supports a full covariance matrix, simplify the problem as typically done in the literature. The simplified version satisfies the following properties:
\begin{itemize}
    \item The mean vectors are placed on the vertices of $Z$-dimensional regular simplex, so that the mean vectors are equidistant. 
    \item The covariance matrices are diagonal, with the same on all the components. In other words, each Gaussian component is \textit{spherical}, formally: $\pmb{\Sigma}_k = \pmb{\sigma}_k\mathbf{I}$, where $\mathbf{I}$ is the identity matrix.
\end{itemize}

\subsection{Implementation of state of the art models}
For all the models but StarGAN*, we used the state of the art implementations released by the authors without any modification. StarGAN* corresponds to a StarGAN model that is conditioned on Gaussian noise ($-0.2 \mathcal{N}(0,1) + 0.1$) in the input domain vector.
We will release the code and trained model of StarGAN*. 

\begin{figure}[!ht]	
	\renewcommand{\tabcolsep}{1pt}
	\renewcommand{\arraystretch}{0.8}
	\newcommand{\sizea}{0.10\linewidth}
	\centering
	\footnotesize
	\begin{tabular}{ccc|ccc|ccc}
	   Image & CAM & Attention & Image & CAM & Attention & Image & CAM & Attention \\ 
		\includegraphics[width=\sizea]{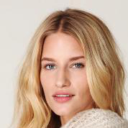} &
		\includegraphics[width=\sizea]{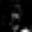} & 
		\includegraphics[width=\sizea]{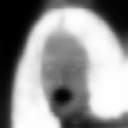}&
		\includegraphics[width=\sizea]{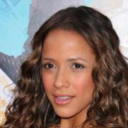} & 
		\includegraphics[width=\sizea]{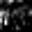} &
		\includegraphics[width=\sizea]{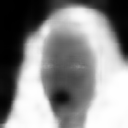} &
		\includegraphics[width=\sizea]{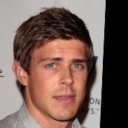} &  \includegraphics[width=\sizea]{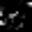} & \includegraphics[width=\sizea]{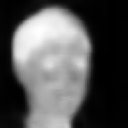} \\
	\end{tabular}
	\caption{Several examples of CAMs and our unsupervised attention masks for hair color translation.}
	\label{Fig:cams}
\end{figure}

\subsection{Class Activation Maps for Faces}

The Faces dataset is a very challenging dataset where each images has multiple attributes, and where the background is very diverse and complex. 
For this reason, we employ an attention mask that helps the network focusing on the attributes that have to be changed. 
However, we found from experimental results that attention is hard to learn. Thus, during the training, we help the network at learning of the unsupervised attention mask through the use Class Activation Maps (CAMs)~\cite{zhou2016learning,selvaraju2017grad}.
We fine-tune the pretrained network VGG-16~\cite{simonyan2014very} on CelebA dataset to do multi-label classification for the selected attributes in the domain translation. 
Then, the predicted one-channel CAM of the real attributes in the original input image is concatenated into the attention layer in decoder $G$. 
Although the CAMs are pretty rough (see Fig.~\ref{Fig:cams}), they improve the unsupervised attention as expected (FID: 48.28 $\to$ FID: 46.21). 
Future work is needed to extend the CAMs method to the multiple attributes settings as in Faces. This would greatly improve the interpretability and efficacy of CAMs.

\section{Additional results}
\subsection{Edges $\leftrightarrow$ shoes: Two-domain translation}
\label{sec:appendix_edgesshoes}
\begin{figure}[ht]
    \setlength{\tabcolsep}{1pt}
	\renewcommand{\arraystretch}{0.8}
    \newcommand{\sizea}{0.13\linewidth}
    \scriptsize
	\centering
	\begin{tabular}{c | ccccc | c}
	Input \& GT & StarGAN* & MUNIT & MSGAN & DRIT++ & GMM-UNIT & BicycleGAN \\ \hline
	   \includegraphics[width=\sizea]{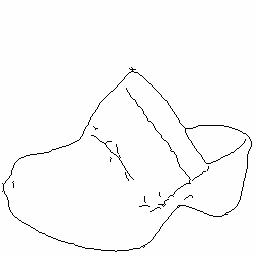} &
       \includegraphics[width=\sizea]{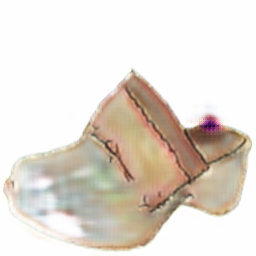} &
	   \includegraphics[width=\sizea]{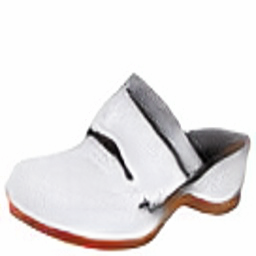} &
	   \includegraphics[width=\sizea]{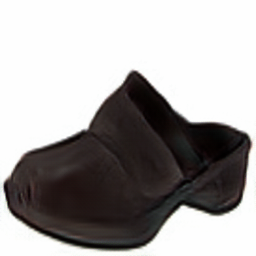} &
	   \includegraphics[width=\sizea]{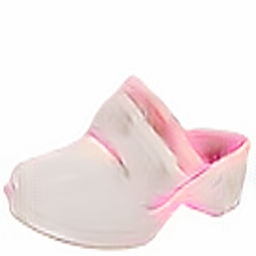} &
	   \includegraphics[width=\sizea]{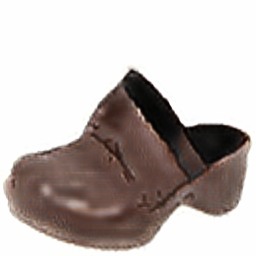} &
	   \includegraphics[width=\sizea]{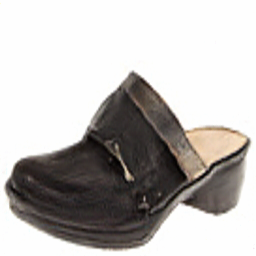}\\ 
	   \includegraphics[width=\sizea]{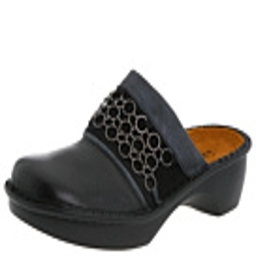} &
	   \includegraphics[width=\sizea]{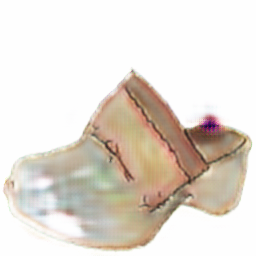} &
	   \includegraphics[width=\sizea]{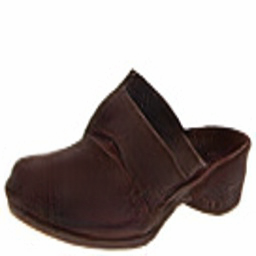} & 
	   \includegraphics[width=\sizea]{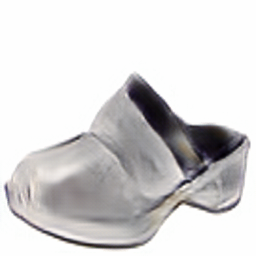} &
	   \includegraphics[width=\sizea]{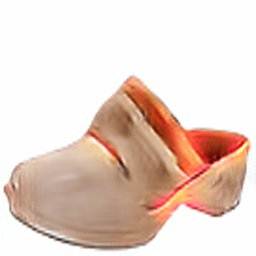} &
	   \includegraphics[width=\sizea]{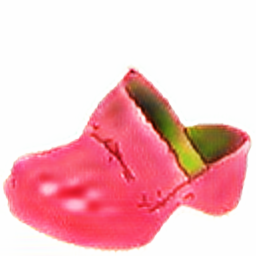} &
	   \includegraphics[width=\sizea]{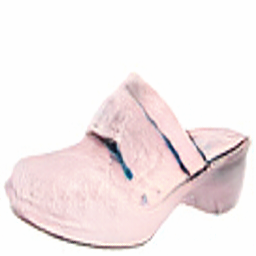}\\ 
	   &
	   \includegraphics[width=\sizea]{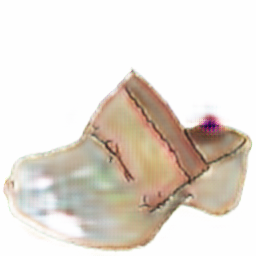} &
	   \includegraphics[width=\sizea]{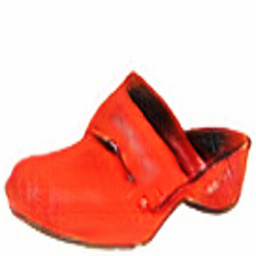} &
	   \includegraphics[width=\sizea]{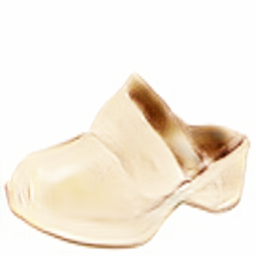} &
	   \includegraphics[width=\sizea]{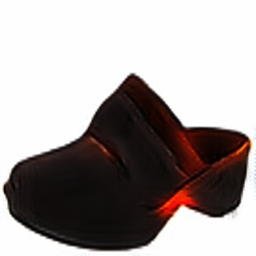} &
	   \includegraphics[width=\sizea]{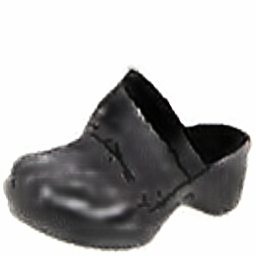} &
	   \includegraphics[width=\sizea]{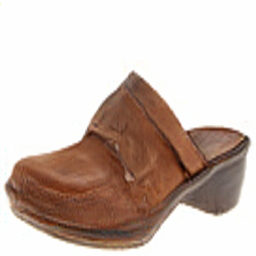}\\ \hline
	   \includegraphics[width=\sizea]{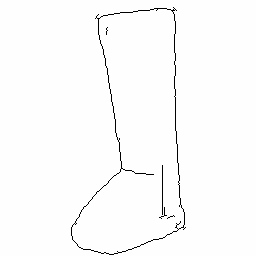} &
       \includegraphics[width=\sizea]{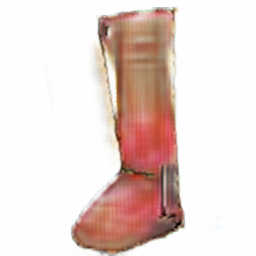} &
	   \includegraphics[width=\sizea]{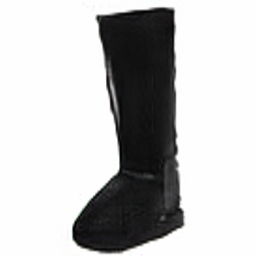} &
	   \includegraphics[width=\sizea]{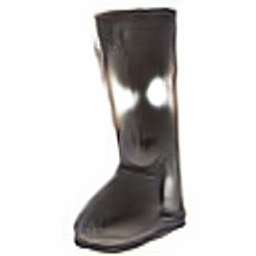} &
	   \includegraphics[width=\sizea]{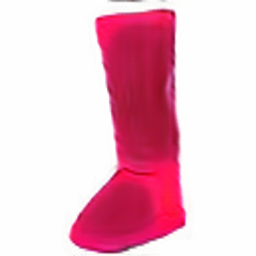} &
	   \includegraphics[width=\sizea]{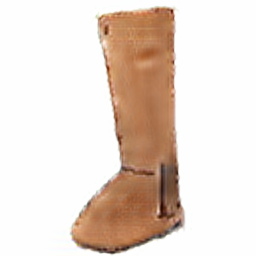} &
	   \includegraphics[width=\sizea]{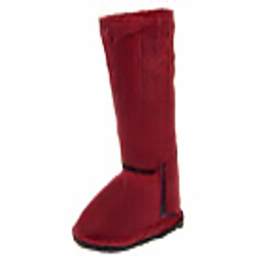}\\ 
	   \includegraphics[width=\sizea]{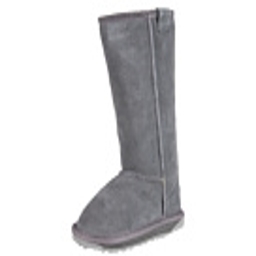} &
	   \includegraphics[width=\sizea]{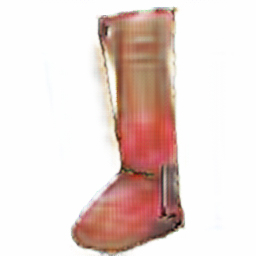} &
	   \includegraphics[width=\sizea]{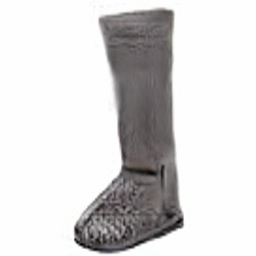} & 
	   \includegraphics[width=\sizea]{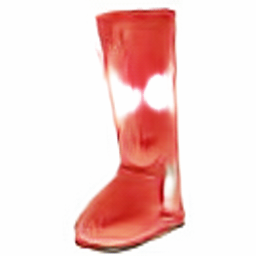} &
	   \includegraphics[width=\sizea]{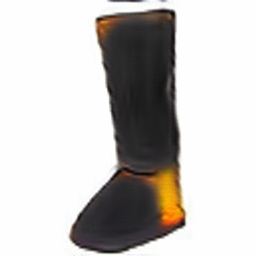} &
	   \includegraphics[width=\sizea]{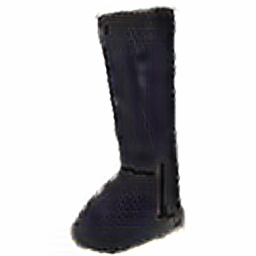} &
	   \includegraphics[width=\sizea]{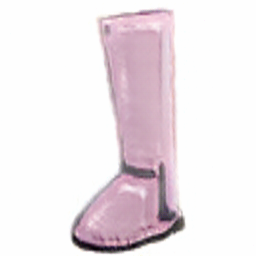}\\ 
	   &
	   \includegraphics[width=\sizea]{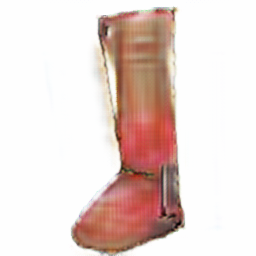} &
	   \includegraphics[width=\sizea]{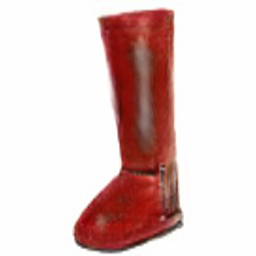} &
	   \includegraphics[width=\sizea]{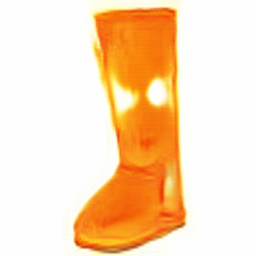} &
	   \includegraphics[width=\sizea]{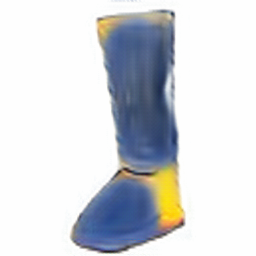} &
	   \includegraphics[width=\sizea]{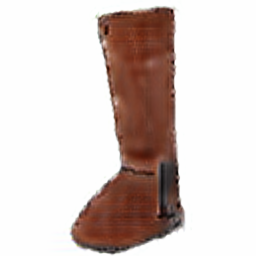} &
	   \includegraphics[width=\sizea]{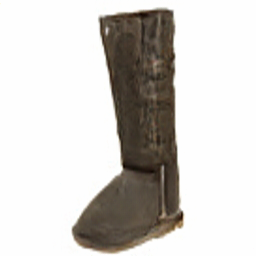}\\ \hline
	   \includegraphics[width=\sizea]{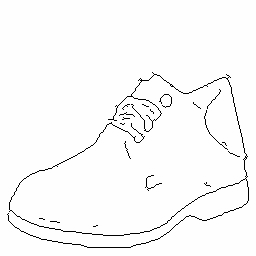} &
       \includegraphics[width=\sizea]{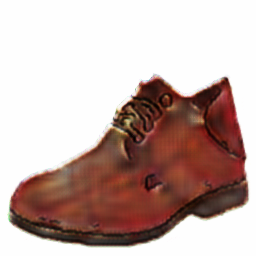} &
	   \includegraphics[width=\sizea]{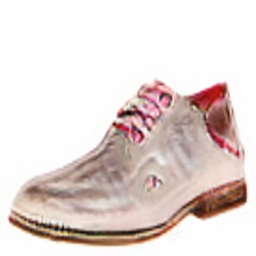} &
	   \includegraphics[width=\sizea]{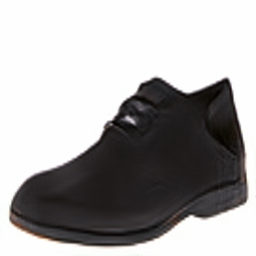} &
	   \includegraphics[width=\sizea]{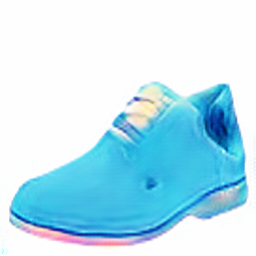} &
	   \includegraphics[width=\sizea]{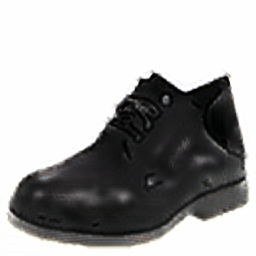} &
	   \includegraphics[width=\sizea]{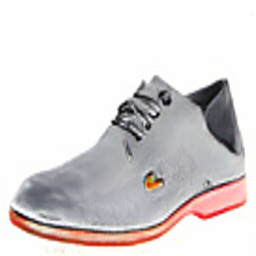}\\
	   \includegraphics[width=\sizea]{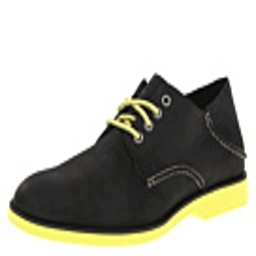} &
	   \includegraphics[width=\sizea]{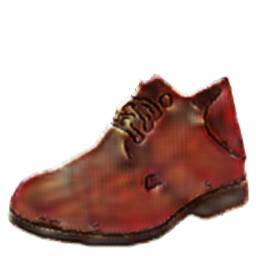} &
	   \includegraphics[width=\sizea]{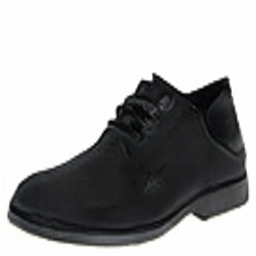} & 
	   \includegraphics[width=\sizea]{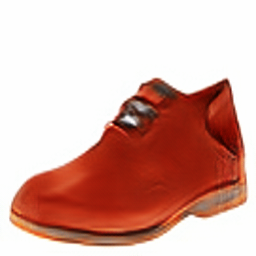} &
	   \includegraphics[width=\sizea]{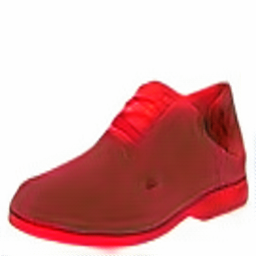} &
	   \includegraphics[width=\sizea]{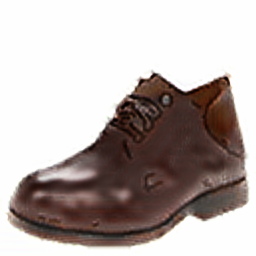} &
	   \includegraphics[width=\sizea]{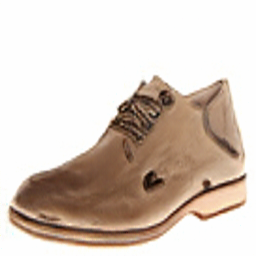}\\
	   &
	   \includegraphics[width=\sizea]{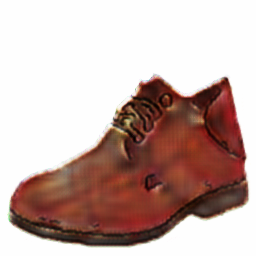} &
	   \includegraphics[width=\sizea]{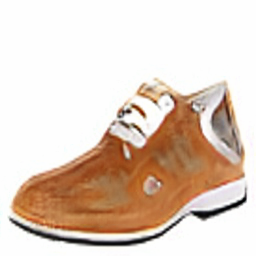} &
	   \includegraphics[width=\sizea]{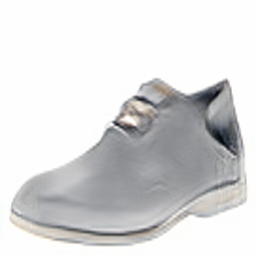} &
	   \includegraphics[width=\sizea]{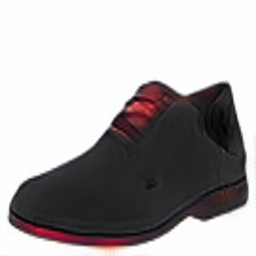} &
	   \includegraphics[width=\sizea]{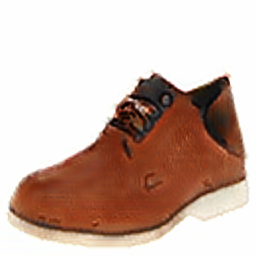} &
	   \includegraphics[width=\sizea]{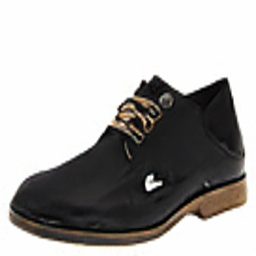}\\ \hline
	\end{tabular}
	\caption{Visual comparisons of state of the art methods on Edge $\leftrightarrow$ Shoes dataset. We note that BicycleGAN, MUNIT and MSGAN are one-to-one domain translation models, while StarGAN* is a multi-domain (deterministic) model. Finally, DRIT++ and GMM-UNIT are multi-modal and multi-domain methods.}
	\label{fig:appendix-edges2shoes}
\end{figure}

In this section, we present the additional results for the one-to-one domain translation. As shown in Figure~\ref{fig:appendix-edges2shoes}, we qualitatively compare GMM-UNIT with the state-of-the-art. We observe that while all the methods (multi-domain and not) achieve acceptable diversity, it seems that DRIT++ suffers from problems of realism. As expected, StarGAN* does not generate diverse results. 

\subsection{Digits: single-attribute multi-domain translation}
\label{sec:appendix_digits}
\begin{figure}[ht]
    \setlength{\tabcolsep}{1pt}
	\renewcommand{\arraystretch}{0.8}
    \newcommand{\sizea}{0.09\linewidth}
    \scriptsize
	\centering
	\begin{tabular}{c ccc  ccc ccc}
	\multirow{2}{*}{Input} & \multicolumn{3}{c}{StarGAN*} & \multicolumn{3}{c}{DRIT++} &  \multicolumn{3}{c}{GMM-UNIT}
	\\
	\cmidrule(lr){2-4} \cmidrule(lr){5-7} \cmidrule(lr){8-10} 
	& MNIST & SVHN & MNISTM & MNIST & SVHN & MNISTM & MNIST & SVHN & MNISTM \\
	   \includegraphics[width=\sizea]{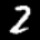} &
       \includegraphics[width=\sizea]{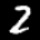} & 
       \includegraphics[width=\sizea]{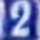} &
	   \includegraphics[width=\sizea]{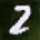} & %
	   \includegraphics[width=\sizea]{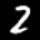} &  
	   \includegraphics[width=\sizea]{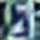} &
	   \includegraphics[width=\sizea]{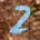} &
	   \includegraphics[width=\sizea]{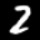} &
	   \includegraphics[width=\sizea]{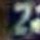} &
	   \includegraphics[width=\sizea]{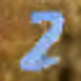} 
	   \\
	   & \includegraphics[width=\sizea]{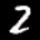} &
	   \includegraphics[width=\sizea]{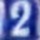} & 
	   \includegraphics[width=\sizea]{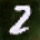} & %
	   \includegraphics[width=\sizea]{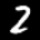} & \includegraphics[width=\sizea]{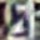} & 
	   \includegraphics[width=\sizea]{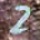} & 
	   \includegraphics[width=\sizea]{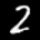} & 
	   \includegraphics[width=\sizea]{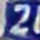} &
	   \includegraphics[width=\sizea]{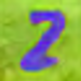} \\
	   & \includegraphics[width=\sizea]{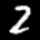}  &
	   \includegraphics[width=\sizea]{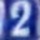} &
	   \includegraphics[width=\sizea]{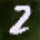} & %
	   \includegraphics[width=\sizea]{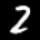} & 
	   \includegraphics[width=\sizea]{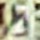} & 
	   \includegraphics[width=\sizea]{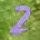} &
	   \includegraphics[width=\sizea]{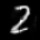} & 
	   \includegraphics[width=\sizea]{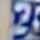} & 
	   \includegraphics[width=\sizea]{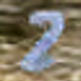} \\
	   \includegraphics[width=\sizea]{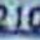} & \includegraphics[width=\sizea]{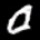}  & 
	   \includegraphics[width=\sizea]{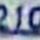}  & 
	   \includegraphics[width=\sizea]{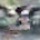}  & %
	   \includegraphics[width=\sizea]{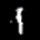} & \includegraphics[width=\sizea]{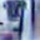} &  
	   \includegraphics[width=\sizea]{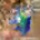} & 
	   \includegraphics[width=\sizea]{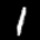} & 
	   \includegraphics[width=\sizea]{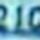} & 
	   \includegraphics[width=\sizea]{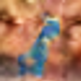}  \\
	   & \includegraphics[width=\sizea]{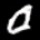} &
	   \includegraphics[width=\sizea]{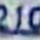} &
	   \includegraphics[width=\sizea]{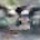}  & %
	   \includegraphics[width=\sizea]{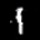} & 
	   \includegraphics[width=\sizea]{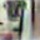} & 
	   \includegraphics[width=\sizea]{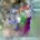} & 
	   \includegraphics[width=\sizea]{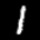} &
	   \includegraphics[width=\sizea]{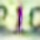} & 
	   \includegraphics[width=\sizea]{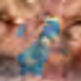} \\
	   & \includegraphics[width=\sizea]{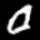} &
	   \includegraphics[width=\sizea]{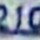} &
	   \includegraphics[width=\sizea]{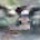}  & %
	   \includegraphics[width=\sizea]{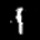} & 
	   \includegraphics[width=\sizea]{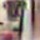} & 
	   \includegraphics[width=\sizea]{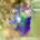} & 
	   \includegraphics[width=\sizea]{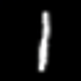} &
	   \includegraphics[width=\sizea]{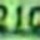} &
	   \includegraphics[width=\sizea]{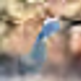}  \\
	   \includegraphics[width=\sizea]{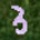} & \includegraphics[width=\sizea]{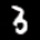} &
	   \includegraphics[width=\sizea]{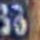} & 
	   \includegraphics[width=\sizea]{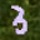} & %
	   \includegraphics[width=\sizea]{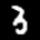} &
	   \includegraphics[width=\sizea]{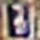} & 
	   \includegraphics[width=\sizea]{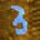} & 
	   \includegraphics[width=\sizea]{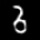} & 
	   \includegraphics[width=\sizea]{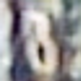} & 
	   \includegraphics[width=\sizea]{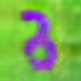}  \\
	   & \includegraphics[width=\sizea]{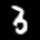} &
	   \includegraphics[width=\sizea]{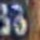} & 
	   \includegraphics[width=\sizea]{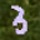} & %
	   \includegraphics[width=\sizea]{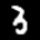} & 
	   \includegraphics[width=\sizea]{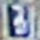} & 
	   \includegraphics[width=\sizea]{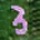} & 
	   \includegraphics[width=\sizea]{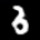} & 
	   \includegraphics[width=\sizea]{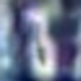} & 
	   \includegraphics[width=\sizea]{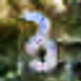}  \\
	   & \includegraphics[width=\sizea]{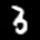}  & 
	   \includegraphics[width=\sizea]{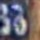} & 
	   \includegraphics[width=\sizea]{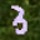} & %
	   \includegraphics[width=\sizea]{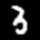} & 
	   \includegraphics[width=\sizea]{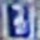} & 
	   \includegraphics[width=\sizea]{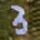} &
	   \includegraphics[width=\sizea]{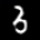} & 
	   \includegraphics[width=\sizea]{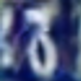} & 
	   \includegraphics[width=\sizea]{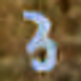} \\
	\end{tabular}
	\caption{Visual comparisons of state of the art methods on the digits dataset. We note that StarGAN* is a multi-domain (deterministic) model, while DRIT++ and GMM-UNIT are multi-modal and multi-domain methods. Image quality is very similar to the input images.}
    \label{fig:appendix-digits}
\end{figure}

\Cref{fig:appendix-digits} shows the qualitative comparison with the state of the art, while \Cref{tab:appendix_QuantitativeResults_digits} show the breakdown, per domain, of the quantitative results. We observe, as expected, that StarGAN* fails at generating diverse results.

\begin{table*}[!ht]
    \small
	\centering
    \setlength{\tabcolsep}{4pt}
	\caption{Quantitative comparison on the  Digits dataset.}
	\begin{tabular}{@{}l lr rr@{}}
	\toprule
		\multirow{2}{*}{\textbf{Target Domain}} &
		\multirow{2}{*}{\textbf{Metric}} &
		\multicolumn{3}{c}{\textbf{Method}}  \\
		\cmidrule{3-5}
		& & StarGAN* & DRIT++&  GMM-UNIT \\
		\midrule
		\multirow{2}{*}{MNIST} & FID$\downarrow$ & 85.11 &  122.59 & $\bst{78.28}$ \\
		& LPIPS$\uparrow$ & 0.002 & 0.001  & $\bst{0.067}$ \\ \midrule
		\multirow{2}{*}{SVHN} & FID$\downarrow$ & 64.91 & 66.88 & $\bst{48.23}$   \\
		& LPIPS$\uparrow$ & 0.006 & 0.045 & $\bst{0.115}$ \\ \midrule
		\multirow{2}{*}{MNIST-M}
		& FID$\downarrow$ & 57.31 & 77.35 & $\bst{55.77}$ \\
		& LPIPS$\uparrow$ & 0.010 & 0.127 & $\bst{0.191}$ \\ \midrule
		& Params$\downarrow$ & 11.18M$\times$1 & 24.49M$\times$1 & 14.26M$\times$1 \\
		\bottomrule
	\end{tabular}
	\label{tab:appendix_QuantitativeResults_digits}
\end{table*}

\subsection{Faces: multi-attribute multi-domain translation}
\label{sec:appendix_faces}
\begin{figure}[ht]
    \setlength{\tabcolsep}{1pt}
    \newcommand{\sizea}{0.13\linewidth}
    \scriptsize
	\centering
	\begin{tabular}{cc ccccc } 
	Input & Model & BA+FM+Y & BN+FM+Y & BW+FM+Y & BN+M+Y & BN+FM+O  \\ \cmidrule{1-7}
	 \includegraphics[width=\sizea]{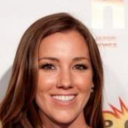} & \multirow{-6}{*}{StarGAN} & 
       \includegraphics[width=\sizea]{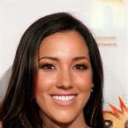} &
	   \includegraphics[width=\sizea]{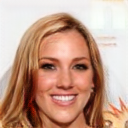} &
	   \includegraphics[width=\sizea]{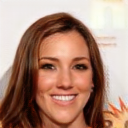} &
	   \includegraphics[width=\sizea]{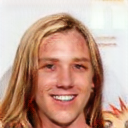} &
	   \includegraphics[width=\sizea]{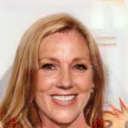} \\ \cmidrule{2-7}
	   & \multirow{8}{*}{GMM-UNIT}  & 
       \includegraphics[width=\sizea]{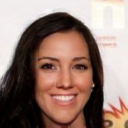} &
       \includegraphics[width=\sizea]{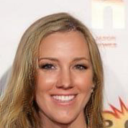} &
       \includegraphics[width=\sizea]{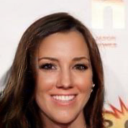} &
       \includegraphics[width=\sizea]{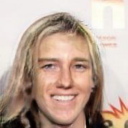} &
       \includegraphics[width=\sizea]{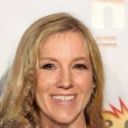} \\ 
	   &   & 
       \includegraphics[width=\sizea]{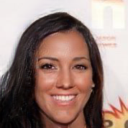} &
       \includegraphics[width=\sizea]{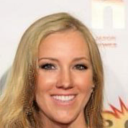} &
       \includegraphics[width=\sizea]{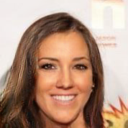} &
       \includegraphics[width=\sizea]{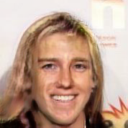} &
       \includegraphics[width=\sizea]{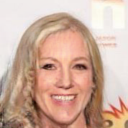} \\ 
	   &   & 
       \includegraphics[width=\sizea]{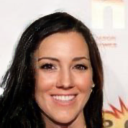} &
       \includegraphics[width=\sizea]{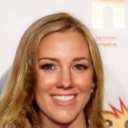} &
       \includegraphics[width=\sizea]{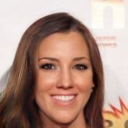} &
       \includegraphics[width=\sizea]{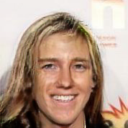} &
       \includegraphics[width=\sizea]{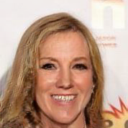} \\ \cmidrule{2-7}
       \includegraphics[width=\sizea]{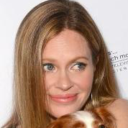} & \multirow{-6}{*}{StarGAN} &
       \includegraphics[width=\sizea]{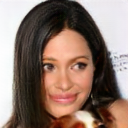} &
	   \includegraphics[width=\sizea]{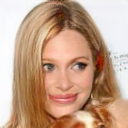} &
	   \includegraphics[width=\sizea]{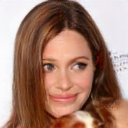} &
	   \includegraphics[width=\sizea]{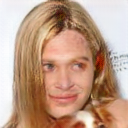} &
	   \includegraphics[width=\sizea]{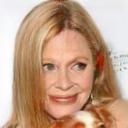} \\ \cmidrule{2-7}
	   & \multirow{8}{*}{GMM-UNIT}  & 
       \includegraphics[width=\sizea]{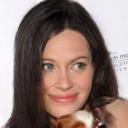} &
       \includegraphics[width=\sizea]{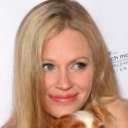} &
       \includegraphics[width=\sizea]{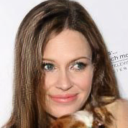} &
       \includegraphics[width=\sizea]{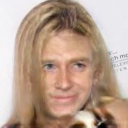} &
       \includegraphics[width=\sizea]{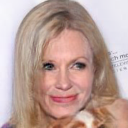} \\ 
	   &   & 
       \includegraphics[width=\sizea]{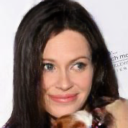} &
       \includegraphics[width=\sizea]{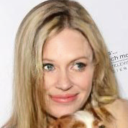} &
       \includegraphics[width=\sizea]{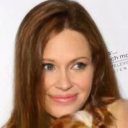} &
       \includegraphics[width=\sizea]{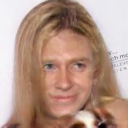} &
       \includegraphics[width=\sizea]{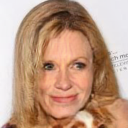} \\ 
	   &   & 
       \includegraphics[width=\sizea]{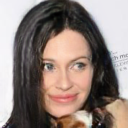} &
       \includegraphics[width=\sizea]{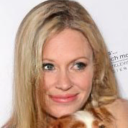} &
       \includegraphics[width=\sizea]{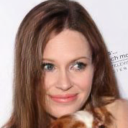} &
       \includegraphics[width=\sizea]{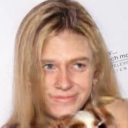} &
       \includegraphics[width=\sizea]{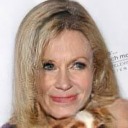} \\ \cmidrule{1-7}
	\end{tabular}
	\caption{Comparisons on CelebA dataset. BA: Black hair, BN: blond hair, BW: Brown hair, M: Male, FM: Female, Y: Young, O: Old.}
	\label{fig:appendix-celeba1}
\end{figure}

\begin{figure}[ht]
    \setlength{\tabcolsep}{1pt}
    \newcommand{\sizea}{0.13\linewidth}
    \scriptsize
	\centering
	\begin{tabular}{cc ccccc } 
	Input & Model & BA+FM+Y & BN+FM+Y & BW+FM+Y & BN+M+Y & BN+FM+O  \\ \cmidrule{1-7}
	 \includegraphics[width=\sizea]{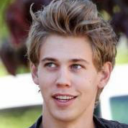} & \multirow{-6}{*}{StarGAN} & 
       \includegraphics[width=\sizea]{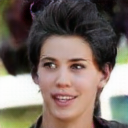} &
	   \includegraphics[width=\sizea]{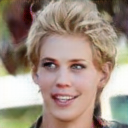} &
	   \includegraphics[width=\sizea]{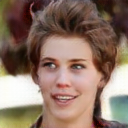} &
	   \includegraphics[width=\sizea]{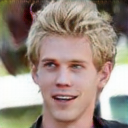} &
	   \includegraphics[width=\sizea]{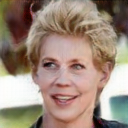} \\ \cmidrule{2-7}
	   & \multirow{8}{*}{GMM-UNIT}  & 
       \includegraphics[width=\sizea]{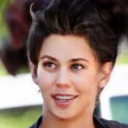} &
       \includegraphics[width=\sizea]{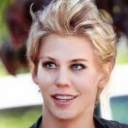} &
       \includegraphics[width=\sizea]{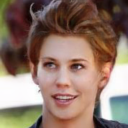} &
       \includegraphics[width=\sizea]{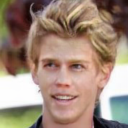} &
       \includegraphics[width=\sizea]{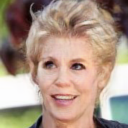} \\ 
	   &   & 
       \includegraphics[width=\sizea]{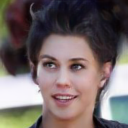} &
       \includegraphics[width=\sizea]{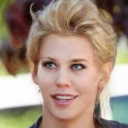} &
       \includegraphics[width=\sizea]{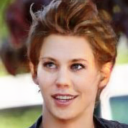} &
       \includegraphics[width=\sizea]{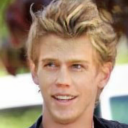} &
       \includegraphics[width=\sizea]{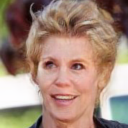} \\ 
	   &   & 
       \includegraphics[width=\sizea]{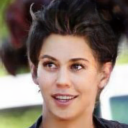} &
       \includegraphics[width=\sizea]{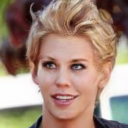} &
       \includegraphics[width=\sizea]{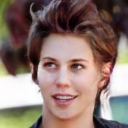} &
       \includegraphics[width=\sizea]{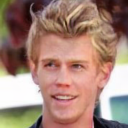} &
       \includegraphics[width=\sizea]{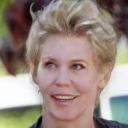} \\ \cmidrule{2-7}
       \includegraphics[width=\sizea]{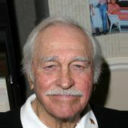} & \multirow{-6}{*}{StarGAN} &
       \includegraphics[width=\sizea]{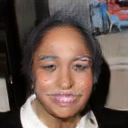} &
	   \includegraphics[width=\sizea]{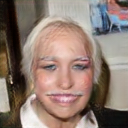} &
	   \includegraphics[width=\sizea]{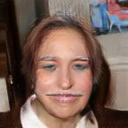} &
	   \includegraphics[width=\sizea]{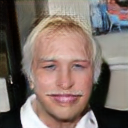} &
	   \includegraphics[width=\sizea]{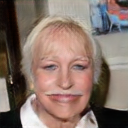} \\ \cmidrule{2-7}
	   & \multirow{8}{*}{GMM-UNIT}  & 
       \includegraphics[width=\sizea]{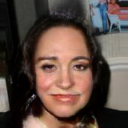} &
       \includegraphics[width=\sizea]{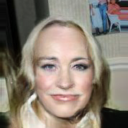} &
       \includegraphics[width=\sizea]{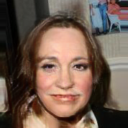} &
       \includegraphics[width=\sizea]{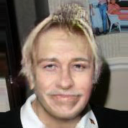} &
       \includegraphics[width=\sizea]{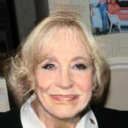} \\ 
	   &   & 
       \includegraphics[width=\sizea]{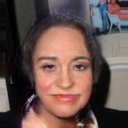} &
       \includegraphics[width=\sizea]{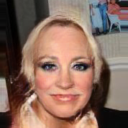} &
       \includegraphics[width=\sizea]{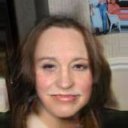} &
       \includegraphics[width=\sizea]{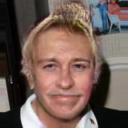} &
       \includegraphics[width=\sizea]{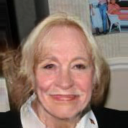} \\ 
	   &   & 
       \includegraphics[width=\sizea]{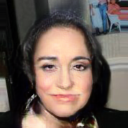} &
       \includegraphics[width=\sizea]{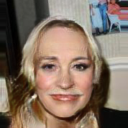} &
       \includegraphics[width=\sizea]{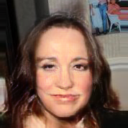} &
       \includegraphics[width=\sizea]{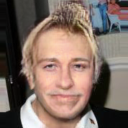} &
       \includegraphics[width=\sizea]{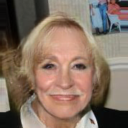} \\ \cmidrule{1-7}
	\end{tabular}
	\caption{Comparisons on CelebA dataset. BA: Black hair, BN: blond hair, BW: Brown hair, M: Male, FM: Female, Y: Young, O: Old.}
	\label{fig:appendix-celeba2}
\end{figure}

\begin{figure*}[!ht]
	\renewcommand{\tabcolsep}{1pt}
	\renewcommand{\arraystretch}{0.8}
	\newcommand{\sizea}{0.13\linewidth}
	\centering
	\begin{tabular}{c|ccccc} 
	\includegraphics[width=\sizea]{figures/use_ref/title.pdf} &%
	\includegraphics[width=\sizea]{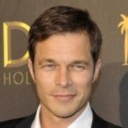} &%
	\includegraphics[width=\sizea]{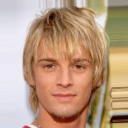} &%
	\includegraphics[width=\sizea]{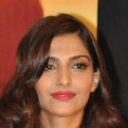} &%
	\includegraphics[width=\sizea]{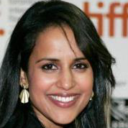} &
	\includegraphics[width=\sizea]{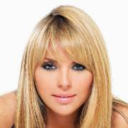} \\
	\includegraphics[width=\sizea]{figures/appendix/style-transfer/000033.png} & 
	\includegraphics[width=\sizea]{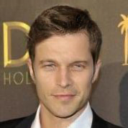} &
	\includegraphics[width=\sizea]{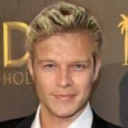} &
	\includegraphics[width=\sizea]{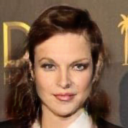} &
	\includegraphics[width=\sizea]{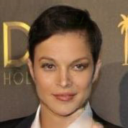} & 
	\includegraphics[width=\sizea]{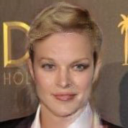}\\
	\includegraphics[width=\sizea]{figures/appendix/style-transfer/055899.png} & 
	\includegraphics[width=\sizea]{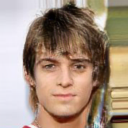} &
	\includegraphics[width=\sizea]{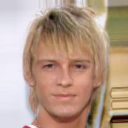} &  
	\includegraphics[width=\sizea]{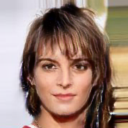} & 
	\includegraphics[width=\sizea]{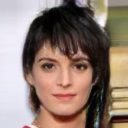} & 
	\includegraphics[width=\sizea]{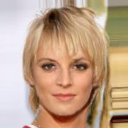}\\
	\includegraphics[width=\sizea]{figures/appendix/style-transfer/000781.png} &
	\includegraphics[width=\sizea]{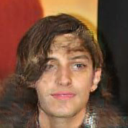} &
	\includegraphics[width=\sizea]{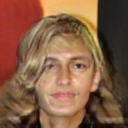}&
	\includegraphics[width=\sizea]{figures/appendix/style-transfer/000781.png} &  
	\includegraphics[width=\sizea]{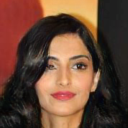} &
	\includegraphics[width=\sizea]{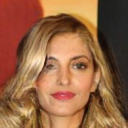} \\
	\end{tabular}
	\caption{Examples of GMM-UNIT applied on the Style transfer task. The style is here extracted from a single reference images provided by the user.}
	\label{Fig:appendix_referebce}
\end{figure*}

\begin{figure*}[!ht]	
	\renewcommand{\tabcolsep}{1pt}
	\renewcommand{\arraystretch}{0.8}
	\newcommand{\sizea}{0.1\linewidth}
	\centering
	\footnotesize
	\begin{tabular}{c|cccccccc}
	        Input & \multicolumn{2}{l}{Blond hair}  & & \multicolumn{2}{c}{$\longleftrightarrow$} & & \multicolumn{2}{r}{Blond hair}\\
		\includegraphics[width=\sizea]{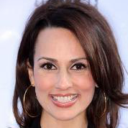} & 
		\includegraphics[width=\sizea]{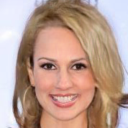} &
		\includegraphics[width=\sizea]{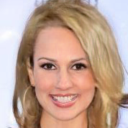} &
		\includegraphics[width=\sizea]{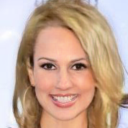} &
		\includegraphics[width=\sizea]{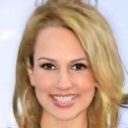} &
		\includegraphics[width=\sizea]{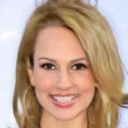} &
		\includegraphics[width=\sizea]{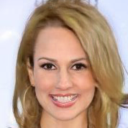} &
		\includegraphics[width=\sizea]{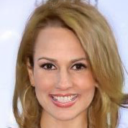} &
		\includegraphics[width=\sizea]{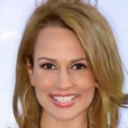} \\
		& \multicolumn{2}{l}{Brown hair} & & \multicolumn{2}{c}{$\longleftrightarrow$} & & \multicolumn{2}{r}{Brown hair}\\
		\includegraphics[width=\sizea]{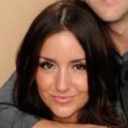} & 
		\includegraphics[width=\sizea]{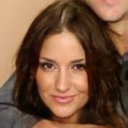} &
		\includegraphics[width=\sizea]{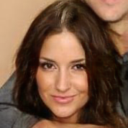} &
		\includegraphics[width=\sizea]{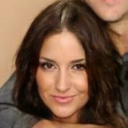} &
		\includegraphics[width=\sizea]{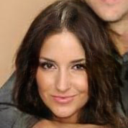} &
		\includegraphics[width=\sizea]{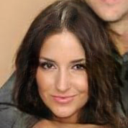} &
		\includegraphics[width=\sizea]{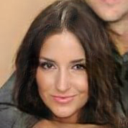} &
		\includegraphics[width=\sizea]{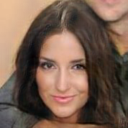} &
		\includegraphics[width=\sizea]{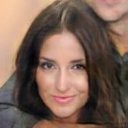} \\
		& \multicolumn{2}{l}{Female} & & \multicolumn{2}{c}{$\longleftrightarrow$} & & \multicolumn{2}{r}{Female}\\
		\includegraphics[width=\sizea]{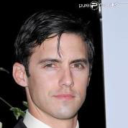} & 
		\includegraphics[width=\sizea]{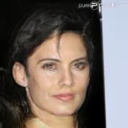} &
		\includegraphics[width=\sizea]{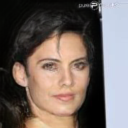} &
		\includegraphics[width=\sizea]{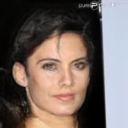} &
		\includegraphics[width=\sizea]{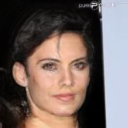} &
		\includegraphics[width=\sizea]{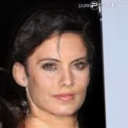} &
		\includegraphics[width=\sizea]{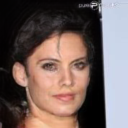} &
		\includegraphics[width=\sizea]{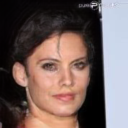} &
		\includegraphics[width=\sizea]{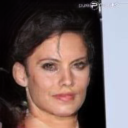} \\
        & \multicolumn{2}{l}{Black hair + Old} & & \multicolumn{2}{c}{$\longleftrightarrow$} & & \multicolumn{2}{r}{Black hair + Old}\\
		\includegraphics[width=\sizea]{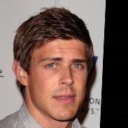} & 
		\includegraphics[width=\sizea]{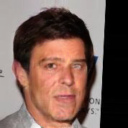} &
		\includegraphics[width=\sizea]{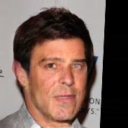} &
		\includegraphics[width=\sizea]{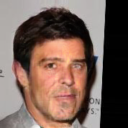} &
		\includegraphics[width=\sizea]{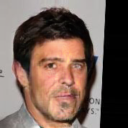} &
		\includegraphics[width=\sizea]{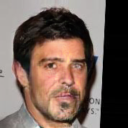} &
		\includegraphics[width=\sizea]{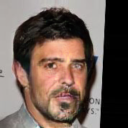} &
		\includegraphics[width=\sizea]{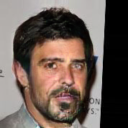} &
		\includegraphics[width=\sizea]{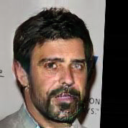} \\
	\end{tabular}
	\caption{Examples of attribute intra-domain interpolation.}
	\label{Fig:intrainterpolation}
\end{figure*}

In \Cref{tab:QuantitativeResults_celeba} we show the quantitative results on the CelebA dataset, broken down per domain. In \Cref{fig:appendix-celeba1} and \Cref{fig:appendix-celeba2}, we show some generated images in comparison with StarGAN. \Cref{Fig:appendix_referebce} shows more examples of manipulating images by using reference images. \Cref{Fig:intrainterpolation} shows the possibility to do attribute interpolation inside a domain, while \Cref{Fig:supplinterpolation} shows the interpolation between domains.

\begin{table*}[!ht]
    \small
	\centering
    \setlength{\tabcolsep}{4pt}
	\caption{Quantitative comparison on the  CelebA dataset.}
	\begin{tabular}{@{}l lr r r@{}}
	\toprule
		\multirow{2}{*}{\textbf{Target Domain}} &
		\multirow{2}{*}{\textbf{Metric}} &
		\multicolumn{3}{c}{\textbf{Method}}  \\
		\cmidrule(r{4pt}){3-5}
		& & StarGAN* & DRIT++ & GMM-UNIT \\
		\midrule
		\multirow{2}{*}{Black hair + Female + Young} & FID$\downarrow$ & 46.80 & 47.94 &  $\bst{39.67}$ \\
		& LPIPS$\uparrow$ & 0.001 & 0.016 & $\bst{0.042}$ \\ \midrule
		\multirow{2}{*}{Blond hair + Female + Young} & FID$\downarrow$ & 63.09 & 71.43 & $\bst{57.38}$  \\
		& LPIPS$\uparrow$ & 0.003 & 0.017 & $\bst{0.060}$ \\ \midrule
		\multirow{2}{*}{Brown hair + Female + Young} & FID$\downarrow$ & 45.15 & 47.54 & $\bst{41.59}$ \\
		& LPIPS$\uparrow$ & 0.003 & 0.017 & $\bst{0.042}$\\  \midrule
	    & Params $\downarrow$ & 53.23M$\times$1 & 54.06M$\times$1 & 26.91M$\times$1 \\
		\bottomrule
	\end{tabular}
	\label{tab:QuantitativeResults_celeba}
\end{table*}

\begin{figure*}[ht]	
	\renewcommand{\tabcolsep}{1pt}
	\renewcommand{\arraystretch}{0.8}
	\newcommand{\sizea}{0.1\linewidth}
	\centering
	\footnotesize
	\begin{tabular}{c|cccccccc}
	        Input & \multicolumn{2}{l}{Black hair} & & \multicolumn{2}{c}{$\longleftrightarrow$} & & \multicolumn{2}{r}{Blond hair}\\
		\includegraphics[width=\sizea]{figures/appendix/interpolation/003453.png} &
		\includegraphics[width=\sizea]{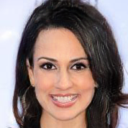} &
		\includegraphics[width=\sizea]{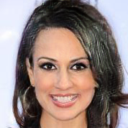} &
        \includegraphics[width=\sizea]{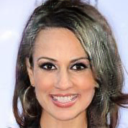} & 
        \includegraphics[width=\sizea]{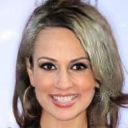} &
        \includegraphics[width=\sizea]{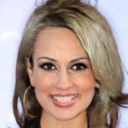} &
        \includegraphics[width=\sizea]{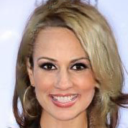} &
        \includegraphics[width=\sizea]{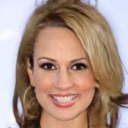} & 
        \includegraphics[width=\sizea]{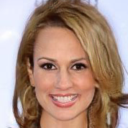} \\
		& \multicolumn{2}{l}{Blond hair+Female} & & \multicolumn{2}{c}{$\longleftrightarrow$} & & \multicolumn{2}{r}{Black hair+Male}\\
		\includegraphics[width=\sizea]{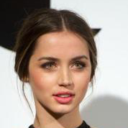} &
		\includegraphics[width=\sizea]{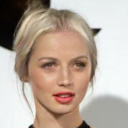} & 
		\includegraphics[width=\sizea]{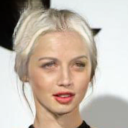} & 
		\includegraphics[width=\sizea]{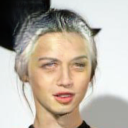} &
		\includegraphics[width=\sizea]{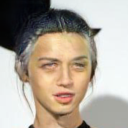} &
		\includegraphics[width=\sizea]{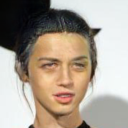} &
		\includegraphics[width=\sizea]{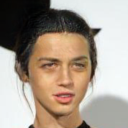} &
		\includegraphics[width=\sizea]{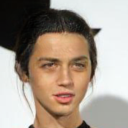} &
		\includegraphics[width=\sizea]{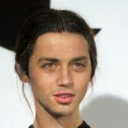} \\
		& \multicolumn{3}{l}{Blond hair+Young}  & \multicolumn{2}{c}{$\longleftrightarrow$} &  \multicolumn{3}{r}{Blond hair+Old}\\
		\includegraphics[width=\sizea]{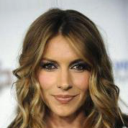} & \includegraphics[width=\sizea]{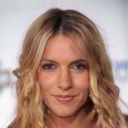} & 
		\includegraphics[width=\sizea]{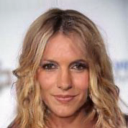} & 
		\includegraphics[width=\sizea]{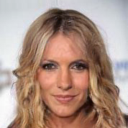} & 
		\includegraphics[width=\sizea]{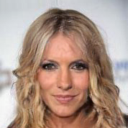} & 
		\includegraphics[width=\sizea]{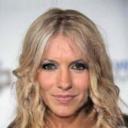} & 
		\includegraphics[width=\sizea]{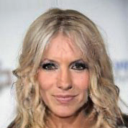} & 
		\includegraphics[width=\sizea]{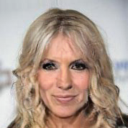} & 
		\includegraphics[width=\sizea]{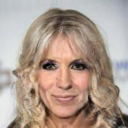} \\
	    & \multicolumn{3}{l}{Blond hair+Young}  & \multicolumn{2}{c}{$\longleftrightarrow$} &  \multicolumn{3}{r}{Brown hair+Old}\\
		\includegraphics[width=\sizea]{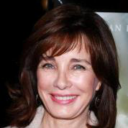} & \includegraphics[width=\sizea]{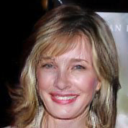} & 
		\includegraphics[width=\sizea]{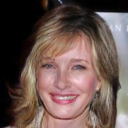} & 
		\includegraphics[width=\sizea]{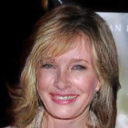} & 
		\includegraphics[width=\sizea]{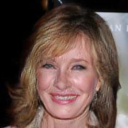} & 
		\includegraphics[width=\sizea]{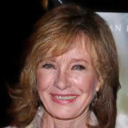} & 
		\includegraphics[width=\sizea]{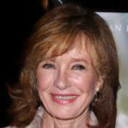} & 
		\includegraphics[width=\sizea]{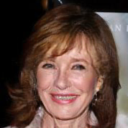} & 
		\includegraphics[width=\sizea]{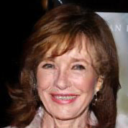} \\
	\end{tabular}
	\caption{Examples of domain interpolation given an input image.}
	\label{Fig:supplinterpolation}
\end{figure*}

\section{Ablation study per domain}
\label{sec:appendix_ablation_digits_details}
In \Cref{tab:appendix_ablation_digits_details} we show additional, per domain, ablation results on the Digits dataset. As it can be seen, we achieve the best image quality results in SVHN and MNISTM but MNIST work better with less complexity. This could be explain by the fact that MNIST is a very simple dataset with only grayscale pixels, where the FID score might be very sensible. 
In all the domain it seems that the network has to achieve a trade-off between quality and diversity, and this trade-off is largely due to $mathcal{L}_{cyc}$. We note that higher diversity can be achieved especially with low-quality images, in which all the pixels can be randomly changed. Thus, the network has to achieve high quality and also diversity in high quality images.

\begin{table*}[!ht]
    \small
	\centering
    \setlength{\tabcolsep}{4pt}
	\caption{Full ablation study performance per domain on the Digits dataset.}
	\begin{tabular}{@{}ll rrrrr@{}}
	\toprule
		\textbf{Target domain} & \textbf{Model} & \textbf{FID$\downarrow$} & \textbf{LPIPS$\uparrow$} \\
		\midrule
	    \multirow{5}{*}{MNIST} &
	     GMM-UNIT w/o $\mathcal{L}_{cyc}$ & 64.92 & 0.066 \\
		& GMM-UNIT w/o $\mathcal{L}_{a/rec}$ & \ 74.32 & 0.059 \\
	     & GMM-UNIT w/o $\mathcal{L}_{iso}$ & \ 77.86 & \ 0.067 \\
		& GMM-UNIT w/o \emph{disent.} & 72.68 & 0.031 \\
		& GMM-UNIT &   78.08 & 0.067  \\ \midrule
		\multirow{5}{*}{SVHN} &
		GMM-UNIT w/o $\mathcal{L}_{cyc}$ & 70.63 & 0.172 \\ 
		& GMM-UNIT w/o $\mathcal{L}_{a/rec}$ & 45.26 & 0.113 \\
		& GMM-UNIT w/o $\mathcal{L}_{iso}$ & 43.33 & 0.110 \\
		& GMM-UNIT w/o \emph{disent.} & 45.97 & 0.092 \\
		& GMM-UNIT & 47.78 & 0.115 \\ \midrule
		\multirow{5}{*}{MNISTM} & 
		 GMM-UNIT w/o $\mathcal{L}_{cyc}$ & 116.64 & 0.162\\
		& GMM-UNIT w/o $\mathcal{L}_{a/rec}$ & 67.01 & 0.189  \\
		& GMM-UNIT w/o $\mathcal{L}_{iso}$ & 69.91 & 0.169 \\
		& GMM-UNIT w/o \emph{disent.} & 63.51 & 0.169 \\
		& GMM-UNIT & 55.44 & 0.191  \\ 
		\bottomrule
	\end{tabular} 
	\label{tab:appendix_ablation_digits_details}
\end{table*}

\section{Visualization of the Attribute Latent space}
\begin{figure*}[!ht]
    \centering
    \includegraphics[width=\linewidth]{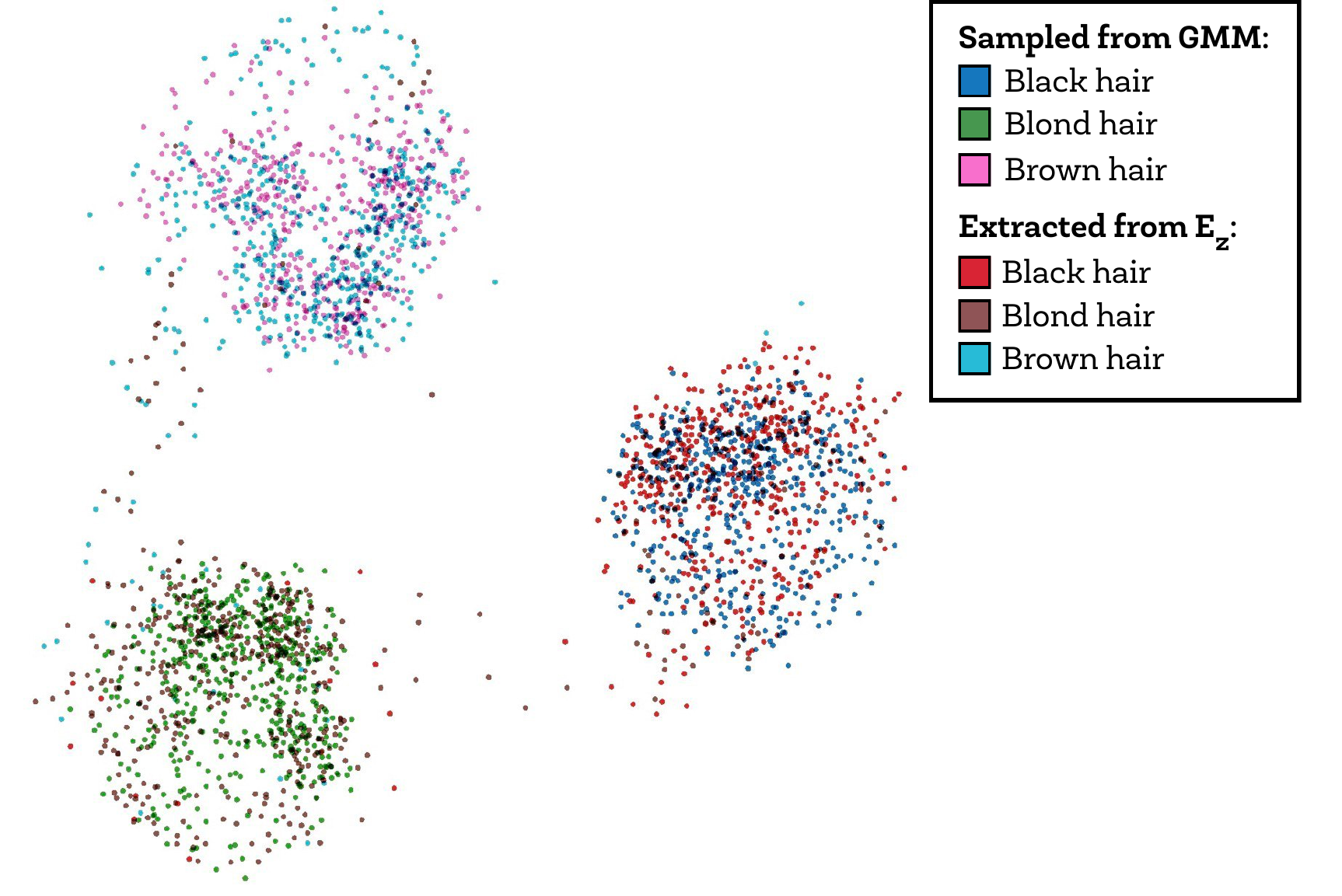}
    \caption{t-SNE projection of the attribute vectors in a 2D space. The points cloud refer to both extracted and sampled attributes, namely black, blond and brown hair, from the GMM-UNIT. The attributes are well separated, while for each attribute the extracted vectors are similar to the sampled ones.}
    \label{fig:projections}
\end{figure*}

In \Cref{fig:projections} we illustrate how three exemplar attributes (black, blond and brown hair) sampled from the GMM distribution are similarly projected in the latent space as those same attributes extracted by the encoder $E_z$. 
To project the attributes to a 2D space we use the t-SNE~\cite{maaten2008visualizing} algorithm with $\text{perplexity}=30, \text{lr}=1.0$ and 300 iterations. 
We can observe from the figure that the attributes are well separated in the space, while the extracted attributes are very close to those sampled. In other words, for example the extracted black hair attribute is most similar to the sampled black hair attribute and most dissimilar to the extracted/sampled attribute of brown hair.

\end{document}